\documentclass{article}

\usepackage{microtype}
\usepackage{graphicx}
\usepackage{subcaption} 
\usepackage{booktabs} 
\usepackage{hyperref}
\usepackage{xspace}
\usepackage{multirow}
\usepackage{enumitem}
\usepackage{bm}
\usepackage{amsmath}
\DeclareMathOperator*{\argmax}{arg\,max}

\usepackage{blindtext}
\usepackage{tikz}
\def\checkmark{\tikz\fill[scale=0.4](0,.35) -- (.25,0) -- (1,.7) -- (.25,.15) -- cycle;}

\usepackage[accepted]{icml2020}

\makeatletter
\renewcommand{\paragraph}{%
 \@startsection{paragraph}{4}%
 {\z@}{0.5ex \@plus 0.5ex \@minus .8ex}{-0.5em}%
 {\normalfont\normalsize\bfseries}%
}
\makeatother

\newcommand{\SIL}{SIL\xspace}
\newcommand{\SILlong}{Seeded Iterated Learning\xspace}

\begin{document}

\twocolumn[
\icmltitle{Countering Language Drift with Seeded Iterated Learning}

\icmlsetsymbol{equal}{*}

\begin{icmlauthorlist}
\icmlauthor{Yuchen Lu}{mila}
\icmlauthor{Soumye Singhal}{mila}
\icmlauthor{Florian Strub}{deepmind}
\icmlauthor{Olivier Pietquin}{brain}
\icmlauthor{Aaron Courville}{mila,cifar}
\end{icmlauthorlist}

\icmlaffiliation{mila}{Mila, University of Montreal}
\icmlaffiliation{cifar}{CIFAR Fellow}
\icmlaffiliation{brain}{Google Research - Brain Team}
\icmlaffiliation{deepmind}{DeepMind}

\icmlcorrespondingauthor{Yuchen Lu}{luyuchen.paul@gmail.com}
\icmlcorrespondingauthor{Soumye Singhal}{singhalsoumye@gmail.com}
\icmlcorrespondingauthor{Florian Strub}{fstrub@google.com}

\icmlkeywords{NLP, Language Drift, Iterated Learning, Interactive Learning}

\vskip 0.3in
]

\printAffiliationsAndNotice{} 
\begin{abstract}
Pretraining on human corpus and then finetuning in a simulator has become a standard pipeline for training a goal-oriented dialogue agent. 
Nevertheless, as soon as the agents are finetuned to maximize task completion, they suffer from the so-called language drift phenomenon: they slowly lose syntactic and semantic properties of language as they only focus on solving the task.
In this paper, we propose a generic approach to counter language drift called \emph{Seeded iterated learning} (SIL). 
We periodically refine a pretrained student agent by imitating data sampled from a newly generated teacher agent. At each time step, the teacher is created by copying the student agent, before being finetuned to maximize task completion.
SIL does not require external syntactic constraint nor semantic knowledge, making it a valuable task-agnostic finetuning protocol. 
We evaluate SIL in a toy-setting Lewis Game, and then scale it up to the translation game with natural language. In both settings, SIL helps counter language drift as well as it improves the task completion compared to baselines.
\end{abstract}
\vspace{-0.2in}

\section{Introduction}
\label{sec:introduction}
Recently, neural language modeling methods have achieved a high level of performance on standard natural language processing tasks~\citep{adiwardana2020towards,radford2019language}. Those agents are trained to capture the statistical properties of language by applying supervised learning techniques over large datasets~\citep{bengio2003neural,collobert2011natural}. While such approaches correctly capture the syntax and semantic components of language, they give rise to inconsistent behaviors in goal-oriented language settings, such as question answering and other dialogue-based tasks~\citep{gao2019neural}. Conversational agents trained via traditional supervised methods tend to output uninformative utterances such as, for example, recommend generic locations while booking for a restaurant~\citep{bordes2016learning}. As models are optimized towards generating grammatically-valid sentences, they fail to correctly ground utterances to task goals ~\citep{strub2017end,lewis2017deal}.


A natural follow-up consists in rewarding the agent to solve the actual language task, rather than solely training it to generate grammatically valid sentences. Ideally, such training would incorporate human interaction~\citep{skantze2010towards,li2016dialogue}, but doing so quickly faces sample-complexity and reproducibility issues. As a consequence, agents are often trained by interacting with a second model to simulate the goal-oriented scenarios~\citep{levin2000stochastic,schatzmann2006survey,lemon2012data}. In the recent literature, a common setting is to pretrain two neural models with supervised learning to acquire the language structure; then, at least one of the agents is finetuned to maximize task-completion with either reinforcement learning, e.g., policy gradient~\citep{williams1992simple}, or Gumbel softmax straight-through estimator~\citep{jang2016categorical,maddison2016concrete}. This finetuning step has shown consistent improvement in dialogue games~\citep{li2016deep,strub2017end,das2017learning}, referential games~\citep{havrylov2017emergence,yu2017joint} or instruction following~\citep{fried2018speaker}.  

\begin{figure*}[th]
    \begin{center}
    \centerline{\includegraphics[width=0.95\linewidth]{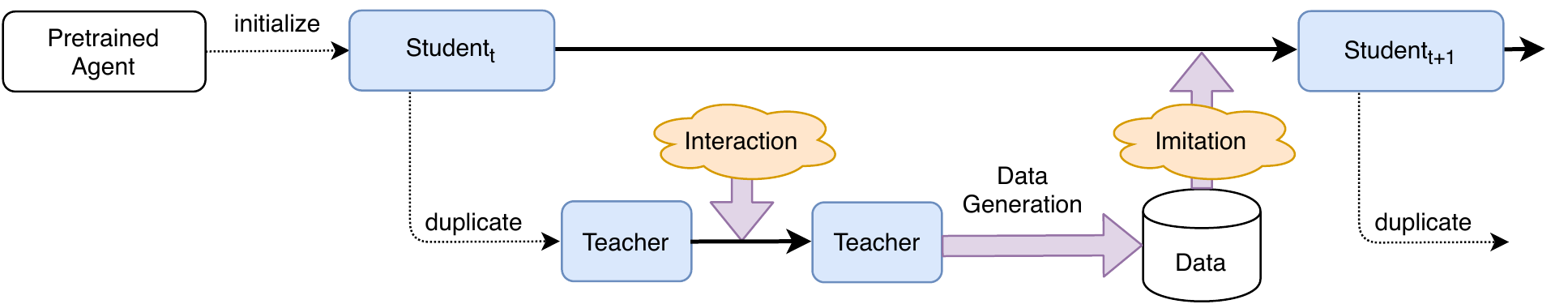}}
    \vskip -0.6em
    \caption{Sketch of \SILlong. 
    A \textbf{student} agent is iteratively refined using newly generated data from a  \textbf{teacher} agent. At each iteration, a teacher agent is created on top of the student before being finetuned by interaction, e.g. maximizing a task completion-score. The teacher then generates a dataset with greedy sampling, which is then used to refine the student through supervised learning. Note that the interaction step involves interaction with another language agent.
    }
    \label{fig:algo_sketch}
    \end{center}
    \vskip -2.3em
\end{figure*}

Unfortunately, interactive learning gives rise to the \emph{language drift} phenomenon. As the agents are solely optimizing for task completion, they have no incentive to preserve the initial language structure.
They start drifting away from the pretrained language output by shaping a task-specific communication protocol. We thus observe a co-adaptation and overspecialization of the agent toward the task, resulting in significant changes to the agent's language distribution.
In practice, there are different forms of language drift~\citep{lazaridou2020multi} including (i) structural drift: removing grammar redundancy (e.g. "is it a cat?" becomes "is cat?"~\citep{strub2017end}), (ii) semantic drift: altering word meaning (e.g. "an old teaching" means "an old man"~\citep{lee2019countering}) or (iii) functional drift: the language results in unexpected actions (e.g. after agreeing on a deal, the agent performs another trade~\citep{li2016deep}). Thus, these agents perform poorly when  paired with humans~\citep{chattopadhyay2017evaluating,zhu2017interactive,lazaridou2020multi}. 

In this paper, we introduce the \emph{\SILlong} (\SIL) protocol to counter language drift. This process is directly inspired by the iterated learning procedure to model the emergence and evolution of language structure~\citep{kirby2001spontaneous,kirby2014iterated}. \SIL does not require human knowledge intervention, it is task-agnostic, and it preserves natural language properties while improving task objectives. 

As illustrated in Figure~\ref{fig:algo_sketch}, \SIL starts from a pretrained agent that instantiates a first generation of \emph{student} agent. The teacher agent starts as a duplicate of the student agent and then goes through a short period of interactive training. Then the teacher generates a training dataset by performing the task over multiple scenarios. Finally, the student is finetuned -- via supervised learning -- to imitate the teacher data, producing the student for next generation, and this process repeats. As further detailed in Section~\ref{sec:itlearn}, the imitation learning step induces a bias toward preserving the well-structured language, while discarding the emergence of specialized and inconsistent language structure~\citep{kirby2001spontaneous}. Finally, \SIL successfully interleaves interactive and supervised learning agents to improves task completions while preserving language properties. 

\textbf{Our contribution} In this work, we propose \SILlong and empirically demonstrate its effectiveness in countering language drift. More precisely, 

\vspace{-0.6em}
\begin{enumerate}[leftmargin=*]
    \setlength\itemsep{0em}  
    \item We study core \SILlong properties on the one-turn Sender-Receiver version of the Lewis Game. 
    \item We demonstrate the practical viability of \SILlong on the French-German translation game that was specifically designed to assess natural language drift~\citep{lee2019countering}. We observe that our method preserves both the semantic and syntactic structure of language, successfully countering language drift while outperforming strong baseline methods. 
    \item We provide empirical evidence towards understanding the algorithm mechanisms\footnote{Code for \href{https://github.com/JACKHAHA363/langauge_drift_lewis_game}{Lewis game} and \href{https://github.com/JACKHAHA363/translation_game_drift}{translation game}}.
\end{enumerate}
\vspace{-1em}

\section{Related Works}
\label{sec:related_works}

        \paragraph{Countering Language Drift} The recent literature on countering language drift includes a few distinct groups of methods.  The first group requires an external labeled dataset, that can be used for visual grounding (i.e. aligning language with visual cues~\citep{lee2019countering}), reward shaping (i.e. incorporating a language metric in the task success score~\citep{li2016deep}) or KL minimization~\citep{havrylov2017emergence}. Yet, these methods depends on the existence of an extra supervision signal and ad-hoc reward engineering, making them less suitable for general tasks. The second group are the population-based methods, which enforces social grounding through a population of agents, preventing them to stray away from the common language~\citep{agarwal2019community}.
        
        The third group of methods involve an alternation between an interactive training phase and a supervised training phase on a pretraining dataset~\citep{wei2018airdialogue, lazaridou2016multi}. This approach has been formalized in \citet{gupta2019seeded} as \emph{Supervised-2-selfPlay} (S2P). Empirically, the S2P approach has shown impressive resistance to language drift and, being relatively task-agnostic, it can be considered a strong baseline for \SIL.
        However, the success of S2P is highly dependent on the quality of the fixed training dataset, which in practice may be noisy, small, and only tangentially related to the task. In comparison, \SIL is less dependent on an initial training dataset since we keep generating new training samples from the teacher throughout training.
        
        \paragraph{Iterated Learning in Emergent Communication}  
        Iterated learning was initially proposed in the field of cognitive science to explore the fundamental mechanisms of language evolution and the persistence of language structure across human generations~\citep{kirby2001spontaneous,kirby2002natural}.
        In particular, \citet{kirby2014iterated} showed that iterated learning consistently turns unstructured proto-language into stable compositional communication protocols in both mathematical modelling and human experiments.
        Recent works~\citep{guo2019emergence, li2019ease, ren2020compositional, cogswell2019emergence, dagan2020co} have extended iterated learning into deep neural networks. They show that the inductive learning bottleneck during the imitation learning phase
        encourages compositionality in the emerged language. Our contribution differs from previous work in this area as we seek to \emph{preserve} the structure of an existing language rather than \emph{emerge} a new structured language.
        
       \paragraph{Lifelong Learning} One of the key problem for neural networks is the problem of catastrophic forgetting~\citep{mccloskey1989catastrophic}. We argue that the problem of language drift can also be viewed as a problem of lifelong learning, since the agent needs to keep the knowledge about language while acquiring new knowledge on using language to solve the task. From this perspective, S2P can be viewed as a method of task rehearsal strategy~\citep{silver2002task} for lifelong learning. The success of iterated learning for language drift could motivate the development of similar methods in countering catastrophic forgetting.

       \paragraph{Self-training}
       Self-training augments the original labeled dataset with unlabeled data paired with the model’s \emph{own} prediction~\citep{he2019revisiting}. After noisy self-training, the student may out-perform the teacher in fields like conditional text generation~\citep{he2019revisiting}, image classification~\citep{xie2019selftraining} and unsupervised machine translation~\citep{lample2017unsupervised}. 
       This process is similar to the imitation learning phase of \SIL except that we only use the self labeled data.

\section{Method}
\label{sec:itlearn}
\paragraph{Learning Bottleneck in Iterated Learning} The core component of iterated learning is the existence of the \emph{learning bottleneck}~\citep{kirby2001spontaneous}: a newly initialized student only acquires the language from a \emph{limited number of examples} generated by the teacher. This bottleneck implicitly favors any structural property of the language that can be exploited by the learner to generalize, such as compositionality. 

Yet, \citet{kirby2001spontaneous} assumes that the student to be a perfect inductive learner that can achieve systematic generalization~\citep{bahdanau2018systematic}. Neural networks are still far from achieving such goal. Instead of using a limited amount of data as suggested, we propose to use a regularization technique, like \emph{limiting the number of imitation steps}, to reduce the ability of the student network to memorize the teacher's data, effectively simulating the learning bottleneck. 
%

%

\paragraph{\SILlong}
\label{sec:sil}

As previously mentioned, \SILlong (\SIL) is an extension of Iterated Learning that aims at preserving an initial language distribution while finetuning the agent to maximize task-score. 
\SIL iteratively refines a pretrained agent, namely the \emph{student}. The \emph{teacher} agent is initially a duplicate of the student agent, and it undergoes an interactive training phase to maximize task score. Then the teacher generates a new training dataset by providing pseudo-labels, and the student performs imitation learning via supervised learning on this synthetic dataset. The final result of the imitation learning will be next student. We repeat the process until the task score converges. The full pipeline is illustrated in Figure~\ref{fig:algo_sketch}.
Methodologically, the key modification of \SIL from the original iterated learning framework is the use of the student agent to seed the imitation learning rather than using a randomly initialized model or a pretrained model.
Our motivation is to ensure a smooth transition during the imitation learning and to retain the task progress.

Although this paper focuses on countering language drift, we emphasize that \SIL is task-agnostic and can be extended to other machine learning settings.

\section{The Sender-Receiver Framework}

    \begin{figure}[t]
      \centering
      \includegraphics[width=1\columnwidth]{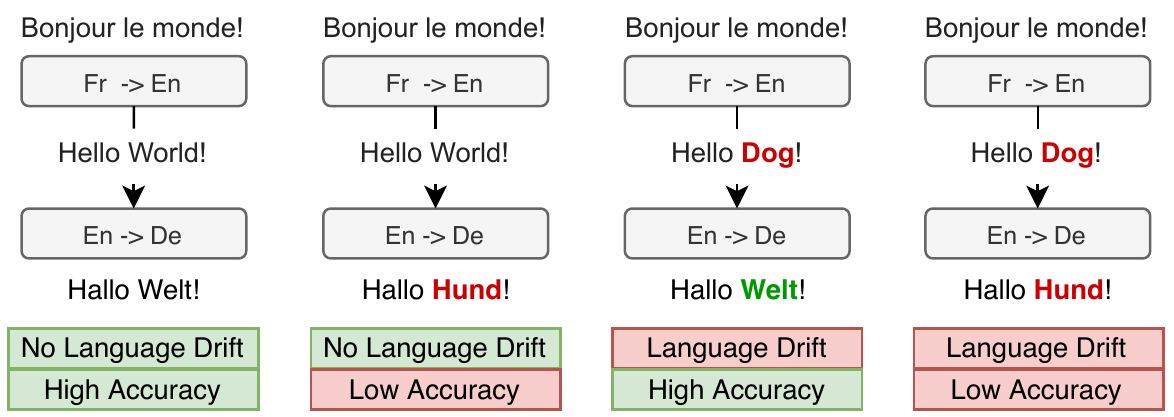}
      \vskip -0.5em
      \caption{In the translation game, the sentence is translated into English then into German. The second and fourth cases are regular failures, while the third case reveals a form of agent co-adaptation.\footnote{Code available at \url{https://github.com/JACKHAHA363/translation_game_drift}}
      }
      \vskip -1em
    \label{fig:translation_game}
    \end{figure}

\begin{figure}[t]
      \centering
      \includegraphics[width=0.8\columnwidth]{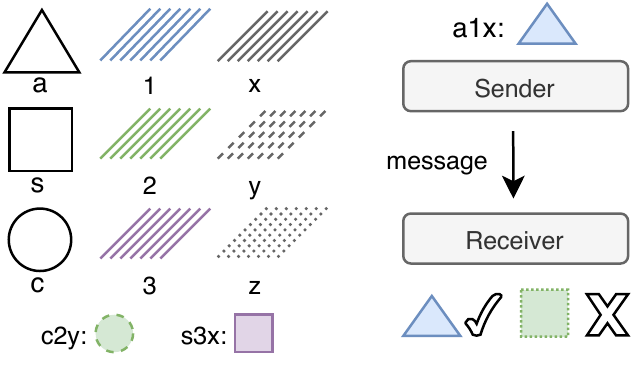}
      \vskip -0.5em
      \caption{\textbf{Lewis game}. Given the input object, the sender emits a compositional message that is parsed by the receiver to retrieve object properties. In the language drift setting, both models are trained toward identity map while solving the reconstruction task.\footnote{Code available at \url{https://github.com/JACKHAHA363/langauge_drift_lewis_game}}}
      \vskip -1em
    \label{fig:lewis_game}
    \end{figure}

We here introduce the experimental framework we use to study the impact of \SIL on language drift. We first introduce the Sender-Receiver (S/R) Game to assess language learning and then detail the instantiation of \SIL for this setting.
\paragraph{Sender-Receiver Games}\label{sec:sr_game} S/R Games are cooperative two-player language games in which the first player, the \emph{sender},  must communicate its knowledge to the second player, the \emph{receiver}, to solve an arbitrary given task. The game can be multi-turn with feedback messages, or single-turn where the sender outputs a single utterance. In this paper, we focus on the single-turn scenario as it eases the language analysis. Yet, our approach may be generalized to multi-turn scenarios. Figures~\ref{fig:translation_game} and~\ref{fig:lewis_game} show two instances of the S/R games studied here: the Translation game~\citep{lee2019countering} and the Lewis game~\citep{kottur2017natural}.

Formally, a single-turn S/R game is defined as a $4$-tuple $\mathcal{G}=(\mathcal{O}, \mathcal{M}, \mathcal{A}, R)$. At the beginning of each episode, an observation (or scenario) $\bm{o}\in\mathcal{O}$ is sampled. Then, the sender $s$ emits a message $\bm{m}=s(\bm{o})\in\mathcal{M}$, where the message can be a sequence of words $m=[w]_{t=1}^{T}$ from a vocabulary $\mathcal{V}$. The receiver $r$ gets the message and performs an action $\bm{a}=r(\bm{m})\in\mathcal{A}$. Finally, both agents receive the same reward $R(\bm{o}, \bm{a})$ which they aim to maximize.

\begin{algorithm}[t]
  \caption{Seeded Iterate Learning for S/R Games}   
  \label{alg:example}
    \small
    \begin{algorithmic}[1]
      \REQUIRE Pretrained parameters of sender $\bm{\theta}$ and receiver $\bm{\phi}$.
      \REQUIRE Training scenarios $\mathcal{O}_{train}$ \hfill \COMMENT{or scenario generator}
      \STATE Copy $\bm{\theta},\bm{\phi}$ to $\bm{\theta}^{S}, \bm{\phi}^{S}$
      \hfill \COMMENT{\emph{Prepare Iterated Learning}}
      \REPEAT
      \STATE Copy $\bm{\theta}^S,\bm{\phi}^S$ to $\bm{\theta}^{T}, \bm{\phi}^{T}$ \hfill \COMMENT{\emph{Initialize Teacher}}
      \FOR{$i=1$ {\bfseries to} $k_1$} 
      \STATE Sample a batch $\bm{o}\in \mathcal{O}_{train}$ 
      \STATE Get $\bm{m} = s(\bm{o};\bm{\theta}^T)$ and $\bm{a} = r(\bm{m};\bm{\phi}^T)$ to have $R(\bm{o},\bm{a})$
      \STATE Update $\bm{\theta}^T$ and $\bm{\phi}^T$ to maximize $R$
      \ENDFOR\hfill \COMMENT{\emph{Finish Interactive Learning}}

      \FOR{$i=1$ {\bfseries to} $k_2$}
      \STATE Sample a batch of $\bm{o}\in \mathcal{O}_{train}$
      \STATE Sample $\bm{m} = s(\bm{o};\bm{\theta}^T)$ 
      \STATE Update $\bm{\theta}^S$ with supervised learning on $(\bm{o}, \bm{m})$
      \ENDFOR \hfill \COMMENT{\emph{Finish Sender Imitation}}
      \FOR{$i=1$ {\bfseries to} $k'_2$}
      \STATE Sample a batch of $\bm{o}\in \mathcal{O}_{train}$
      \STATE Get $\bm{m} = s(\bm{o};\bm{\theta}^S)$ and $\bm{a} = r(\bm{m};\bm{\phi}^S)$ to have $R(\bm{o},\bm{a})$
      \STATE Update $\bm{\phi}^S$ to maximize $R$
      \ENDFOR\hfill \COMMENT{\emph{Finish Receiver Finetuning}}
      \UNTIL{Convergence or maximum steps reached}
    \end{algorithmic}
\end{algorithm}

\paragraph{\SIL For S/R Game} We consider two parametric models, the sender $s(.;\bm{\theta})$ and the receiver $r(.;\bm{\phi})$. Following the \SIL pipeline, we use the uppercase script $S$ and $T$ to respectively denote the parameters of the student and teacher. For instance, $r(.;\bm{\phi}^T)$ refers to the teacher receiver. We also assume that we have a set of scenarios $\mathcal{O}_{train}$ that are fixed or generated on the fly. We detail the \SIL protocol for single-turn S/R games in Algorithm~\ref{alg:example}.

In one-turn S/R games, the language is only emitted by the sender while the receiver's role is to interpret the sender's message and use it to perform the remaining task. 
With this in mind, we train the sender through the \SIL pipeline as defined in Section~\ref{sec:sil} (i.e., interaction, generation, imitation), while we train the receiver to quickly adapt to the new sender's language distribution with a goal of stabilizing training~\citep{ren2020compositional}. First, we jointly train $s(.;\bm{\phi}^T)$ and $r(.;\bm{\phi}^T)$ during the \SIL interactive learning phase.
Second, the sender student imitates the labels generated by $s(.;\bm{\phi}^T)$ through greedy sampling.
Third, the receiver student is trained by maximizing the task score $R(r(\bm{m};\bm{\phi}^S), \bm{o})$ where $\bm{m}=s(\bm{o};\bm{\theta}^S)$ and $\bm{o}\in\mathcal{O}_{train}$. In other words, we finetune the receiver with interactive learning while freezing the new sender parameters. \SIL has three training hyperparameters: (i) $k_1$, the number of interactive learning steps that are performed to obtain the teacher agents, (ii) $k_2$, the number of sender imitation steps, (iii) $k'_2$, the number of interactive steps that are performed to finetune the receiver with the new sender. Unless stated otherwise, we define $k_2 = k'_2$.

\paragraph{Gumbel Straight-Through Estimator}

    In the one-turn S/R game, the task success can generally be described as a differentiable loss such as cross-entropy to update the receiver parameters. Therefore, we here assume that the receiver $r$ can maximize task-completion by minimizing classification or regression errors.
    To estimate the task loss gradient with respect to the sender $s$ parameters, the receiver gradient can be further backpropagated using the Gumbel softmax straight-through estimator (GSTE)~\citep{jang2016categorical,maddison2016concrete}. Hence, the sender parameters are directly optimized toward task loss. Given a sequential message $\bm{m}=[w]_{t=1}^T$, we define $\bm{y}_t$ as follows:
    \begin{equation}
        \bm{y}_t = \mathrm{softmax}\big((\log s(w|\bm{o}, w_{t-1}, \cdots, w_0; \theta) + g_t)/\tau\big)
    \end{equation}
    where $s(w|\bm{o},w_{t-1}, \cdots, w_0)$ is the categorical probability of next word given the sender observation $o$ and previously generated tokens, $g_t \sim \mathrm{Gumbel}(0,1)$ and $\tau$ is the Gumbel temperature that levels exploration. When not stated otherwise, we set $\tau=1$.
    Finally, we sample the next word by taking $w_t = \argmax \bm{y}_t$ before using the straight-through gradient estimator to approximate the sender gradient:
    \begin{equation}
        \frac{\partial R}{\partial \theta} = \frac{\partial R}{\partial w_t}\frac{\partial w_t}{\partial y_t}\frac{\partial y_t}{\partial \theta}\approx \frac{\partial R}{\partial w_t}\frac{\partial y_t}{\partial \theta}.
    \end{equation}
    \SIL can be applied with RL methods when dealing with non-differential reward metrics~\citep{lee2019countering}, however RL  has high gradient variance and we want to GSTE as a start. 
    Since GSTE only optimizes for task completion, language drift will also appear.

\section{Building Intuition: The Lewis Game}

    In this section, we explore a toy-referential game based on the Lewis Game~\citep{lewis1969convention} to have a fine-grained analysis of language drift while exploring the impact of \SIL.

    \begin{figure}[t]
        \begin{subfigure}[b]{0.5\columnwidth}
          \centering
          \includegraphics[width=1\linewidth]{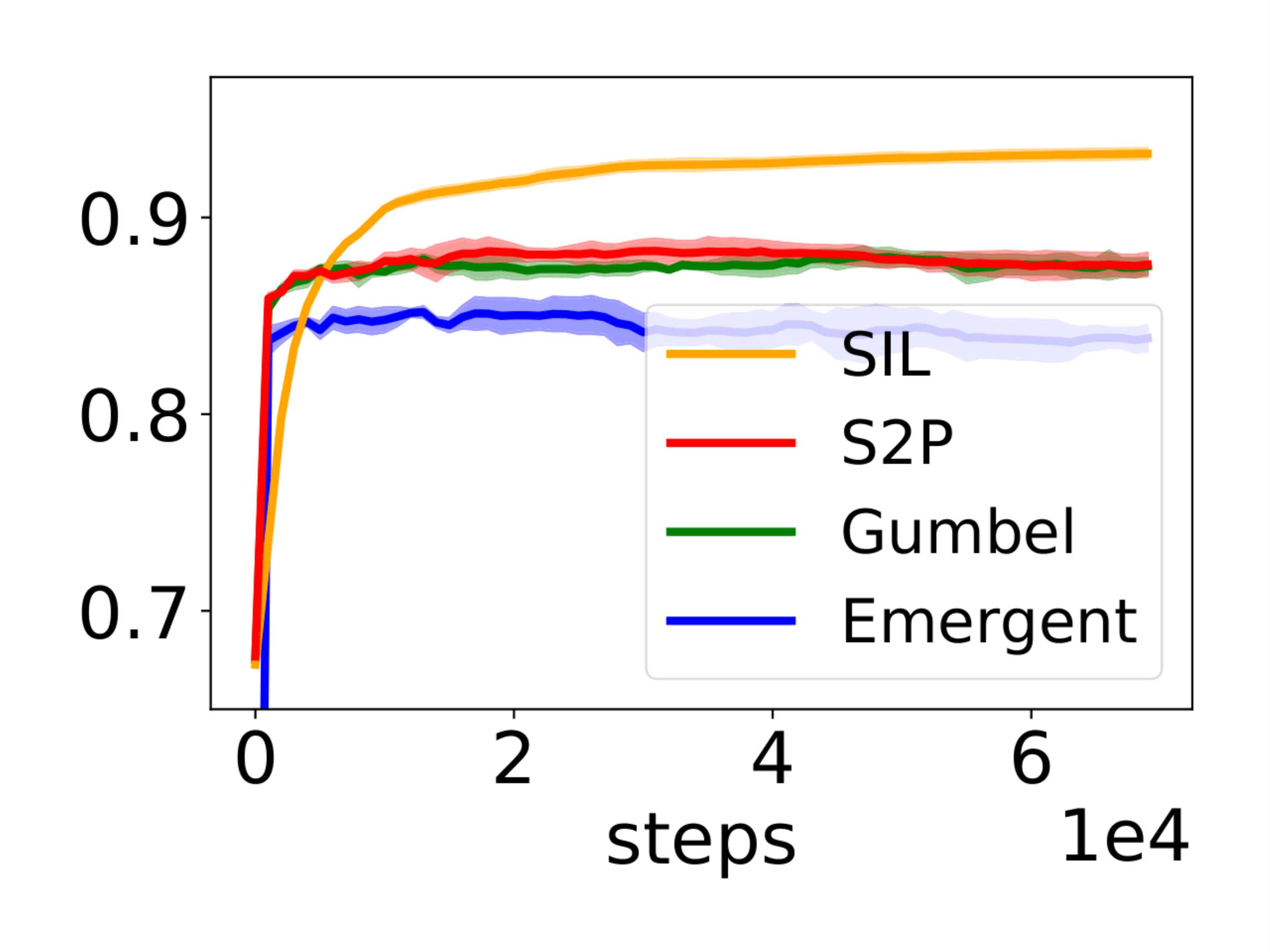}
          \vskip -0.7em
          \caption{Task Score}
        \end{subfigure}
        \hspace{-0.5em}%
        \begin{subfigure}[b]{0.5\columnwidth}
          \centering
          \includegraphics[width=1\linewidth]{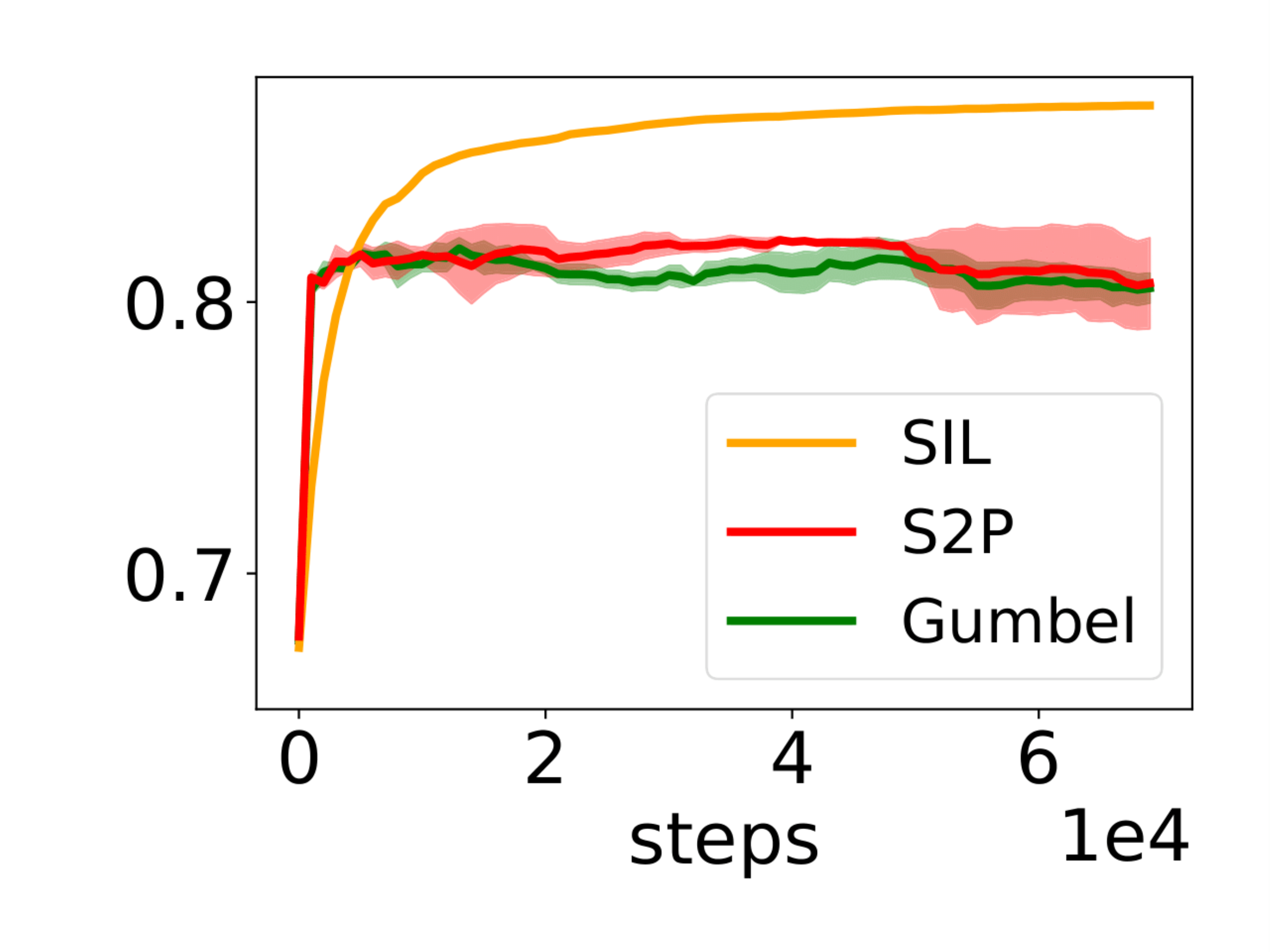}
          \vskip -0.7em
          \caption{Sender Language Score}
        \end{subfigure}
        \vskip -0.7em
        \caption{Task Score and Language Score for \SIL($\tau=10)$ vs baselines ($\tau=1$). \SIL clearly outperforms the baselines. For \SIL: $k_1=1000, k_2=k'_2=400$. The emergent language score is close to zero. All results are averaged over four seeds.}
        \vskip -1em
        \label{fig:lewis_curve}
    \end{figure}

    \paragraph{Experimental Setting} 
    We summarize the Lewis game instantiation described in \citet{gupta2019seeded} to study language drift, and we illustrate it in Figure~\ref{fig:lewis_game}. First, the sender observes an object $o$ with $p$ properties and each property has $t$ possible values: $o[i]\in[1\dots t]$ for $i\in [1\dots p]$. The sender then sends a message $m$ of length $p$ from the vocabulary of size $p\times t$, equal to the number of property values. Our predefined language $\mathcal{L}$ uniquely map each property value to each word, and the message is defined as $\mathcal{L}(o)=[o_1, t + o_2,..., (p-1)t + o_p]$. 
    We study whether this language mapping is preserved during S/R training.
    
    The sender and receiver are modeled by two-layer feed-forward networks. In our task, we use $p=t=5$ with a total of 3125 unique objects. We split this set of objects into three parts: the first split(pre-train) is labeled with correct messages to pre-train the initial agents. The second split is used for the training scenarios. The third split is held out (HO) for final evaluation. The dataset split and hyper-parameters can be found in the Appendix~\ref{sec:appendix_lewis_hyper}. 
    
    We use two main metrics to monitor our training: \emph{Sender Language Score} (LS) and \emph{Task Score} (TS). 
    For the sender language score, we enumerate the held-out objects and compare the generated messages with the ground-truth language on a per token basis. For task accuracy, we compare the reconstructed object vs. the ground-truth object for each property. Formally, we have:
        \vspace{-0.5em}
    \begin{equation}
               LS = \frac{1}{|\mathcal{O}_{HO}|p}\sum_{o\in \mathcal{O}_{HO}}\sum_{l=1}^p [\mathcal{L}(o)[l] == s(o)[l]],
    \end{equation}
    \vspace{-0.5em}
    \begin{equation}
       TS = \frac{1}{|\mathcal{O}_{HO}|p}\sum_{o\in \mathcal{O}_{HO}}\sum_{l=1}^p [o[l] == r(s(o))[l]].
    \end{equation}
    where $[\cdot]$ is the Iverson bracket.
    \paragraph{Baselines} In our experiments, we compare \SIL with different baselines. All methods are initialized with the same pretrained model unless stated otherwise. The \emph{Gumbel} baselines are finetuned with GSTE during interaction. These correspond to naive application of interactive training and are expected to exhibit language drift. 
    \emph{Emergent} is a random initializion trained with GSTE. \emph{S2P} indicates that the agents are trained with Supervised-2-selfPlay. Our \emph{S2P} is realized by using a weighted sum of the losses at each step:
    $
        L_{S2P} = L_{Gumbel} + \alpha L_{supervised}
    $
    where $L_{supervised}$ is the loss on the pre-train dataset and $\alpha$ is a hyperparameter with a default value of 1 as detailed in~\citep{lazaridou2016multi,lazaridou2020multi}.

        \begin{figure}[t]
        \centering
        \begin{subfigure}[b]{0.32\columnwidth}
          \includegraphics[width=\linewidth]{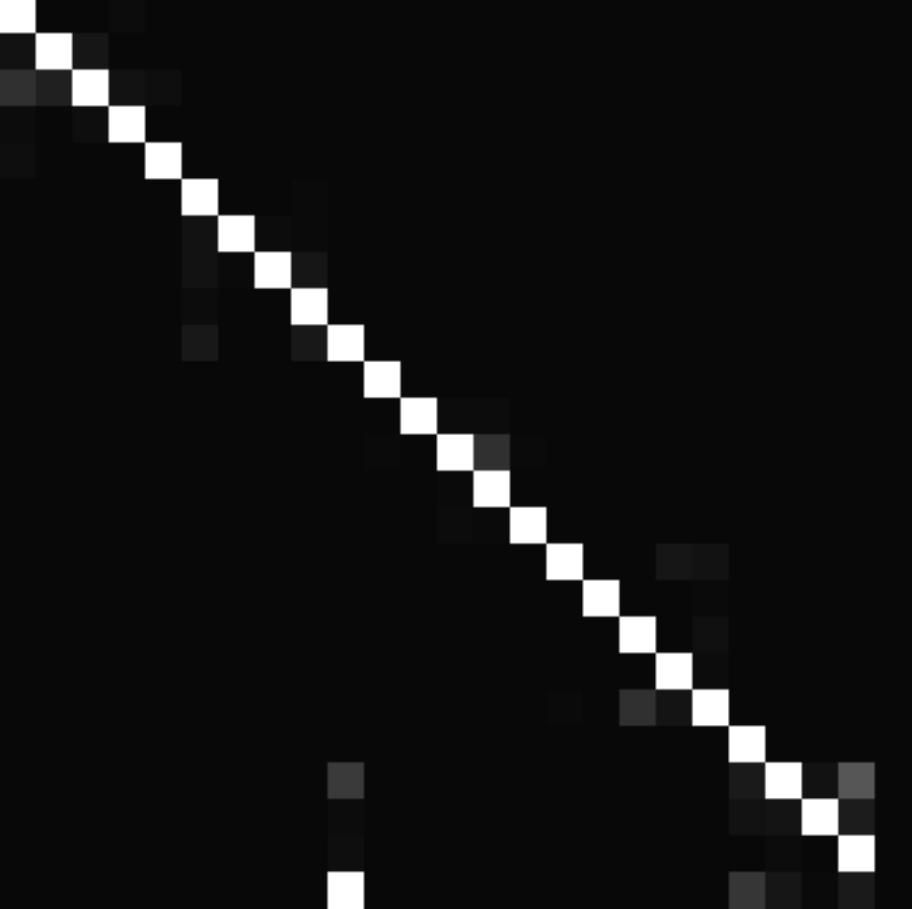}  
          \caption{\SIL}
        \end{subfigure}
        \hspace{0.05em}%
        \begin{subfigure}[b]{0.32\columnwidth}
          \includegraphics[width=\linewidth]{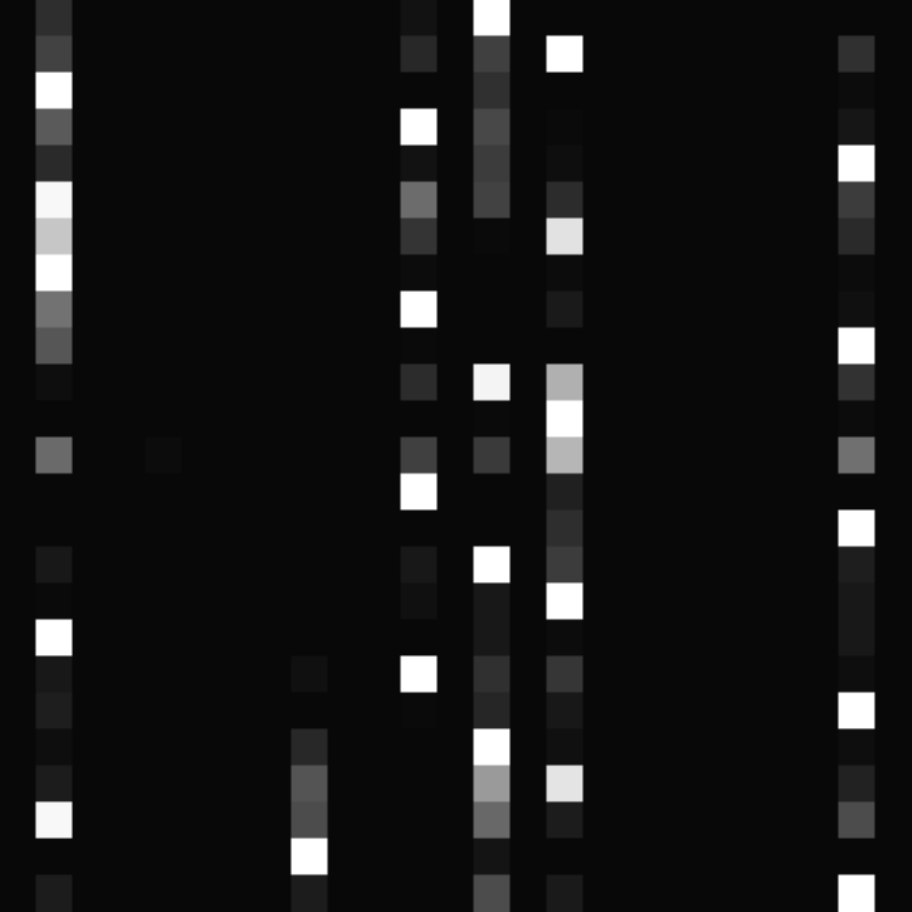}  
          \caption{Emergent}
        \end{subfigure}
        \hspace{0.05em}%
        \begin{subfigure}[b]{0.32\columnwidth}
          \includegraphics[width=\linewidth]{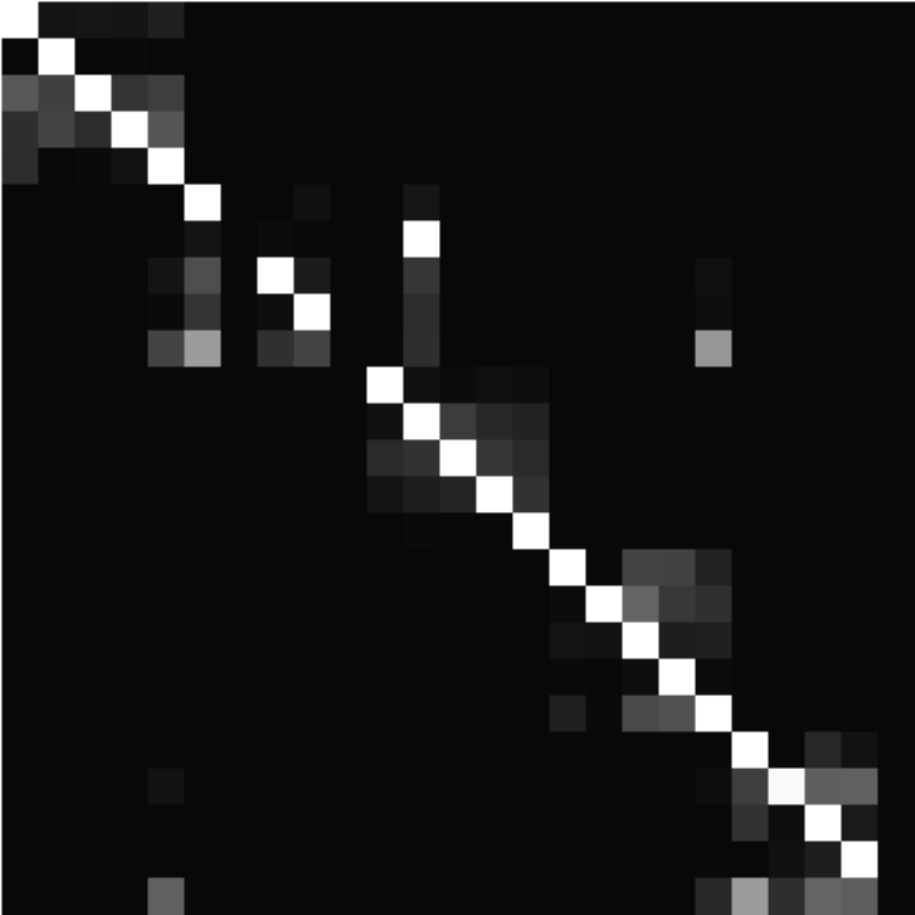}  
          \caption{Gumbel}
        \end{subfigure}
        \vskip -0.5em
        \caption{Comparison of sender's map, where the columns are words and rows are property values. Emergent communication uses the same word to refer to multiple property values.
        A perfect mapped language would be the identity matrix.}
        \vskip -0.5em
        \label{fig:lewis_confusion_matrix}
    \end{figure}
     \paragraph{Results} We present the main results for the Lewis game in Figure~\ref{fig:lewis_curve}. For each method we used optimal hyperparameters namely $\tau = 10$ for \SIL and $\tau=1$ for rest. We also observed that \SIL outperforms the baselines for any $\tau$. Additional results in Appendix~\ref{sec:appendix_lewis} (Figures~\ref{fig:full_lewis_curve_10} \&~\ref{fig:full_lewis_curve_1}). 

     The pretrained agent has an initial task score and language score of around 65\%, showing an imperfect language mapping while allowing room for task improvement. 
     Both Gumbel and S2P are able to increase the task and language score on the held-out dataset. For both baselines, the final task score is higher than the language score. This means that some objects are reconstructed successfully with incorrect messages, suggesting language drift has occurred.
     
     Note that, for S2P, there is some instability of the language score at the end of training. We hypothesize that it could be because our pretrained dataset in this toy setting is too small, and as a result, S2P overfits that small dataset.
     Emergent communication has a sender language score close to zero, which is expected. 
     However, it is interesting to find that emergent communication has slightly lower held-out task score than Gumbel, suggesting that starting from pretrained model provides some prior for the model to generalize better. 
     Finally, we observe that \SIL achieves a significantly higher task score and sender language score, outperforming the other baselines. 
     A high language score also shows that the sender leverages the initial language structure rather than merely re-inventing a new language, countering language drift in this synthetic experiment.  

     To better visualize the underlying language drift in this settings, we display the sender's map from property values to words in Figure \ref{fig:lewis_confusion_matrix}.
     We observe that the freely emerged language results in re-using the same words for different property values. If the method has a higher language score, the resulting map is closer to the identity matrix. 
     
     
    \begin{figure}[t]  

      \begin{subfigure}[b]{0.48\linewidth}
          \centering
          \includegraphics[width=1.0\linewidth]{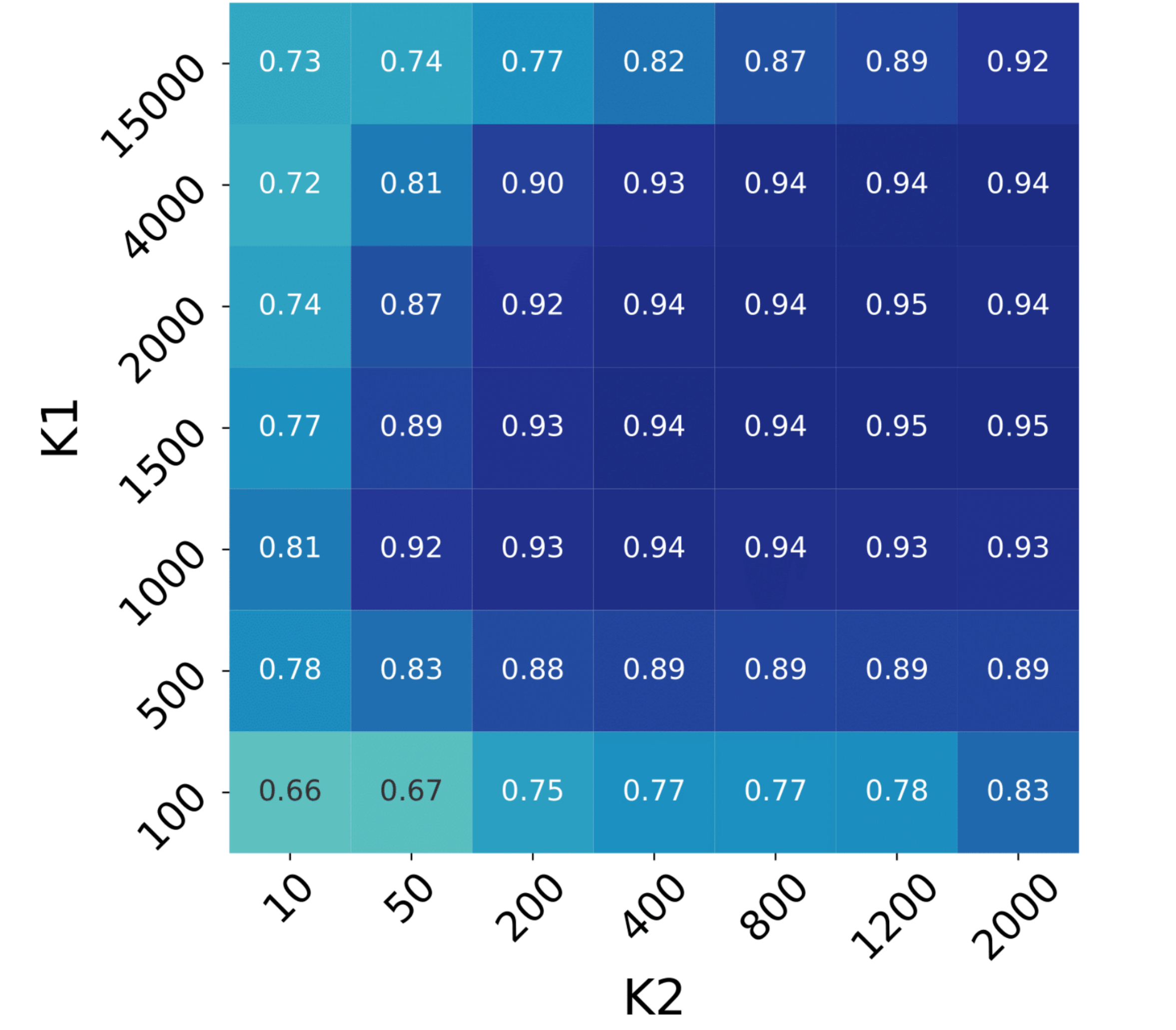}
          \vskip -0.2em
          \caption{\small{Task Score}}
      \end{subfigure}
      \begin{subfigure}[b]{0.48\linewidth}
          \centering
          \includegraphics[width=1.0\linewidth]{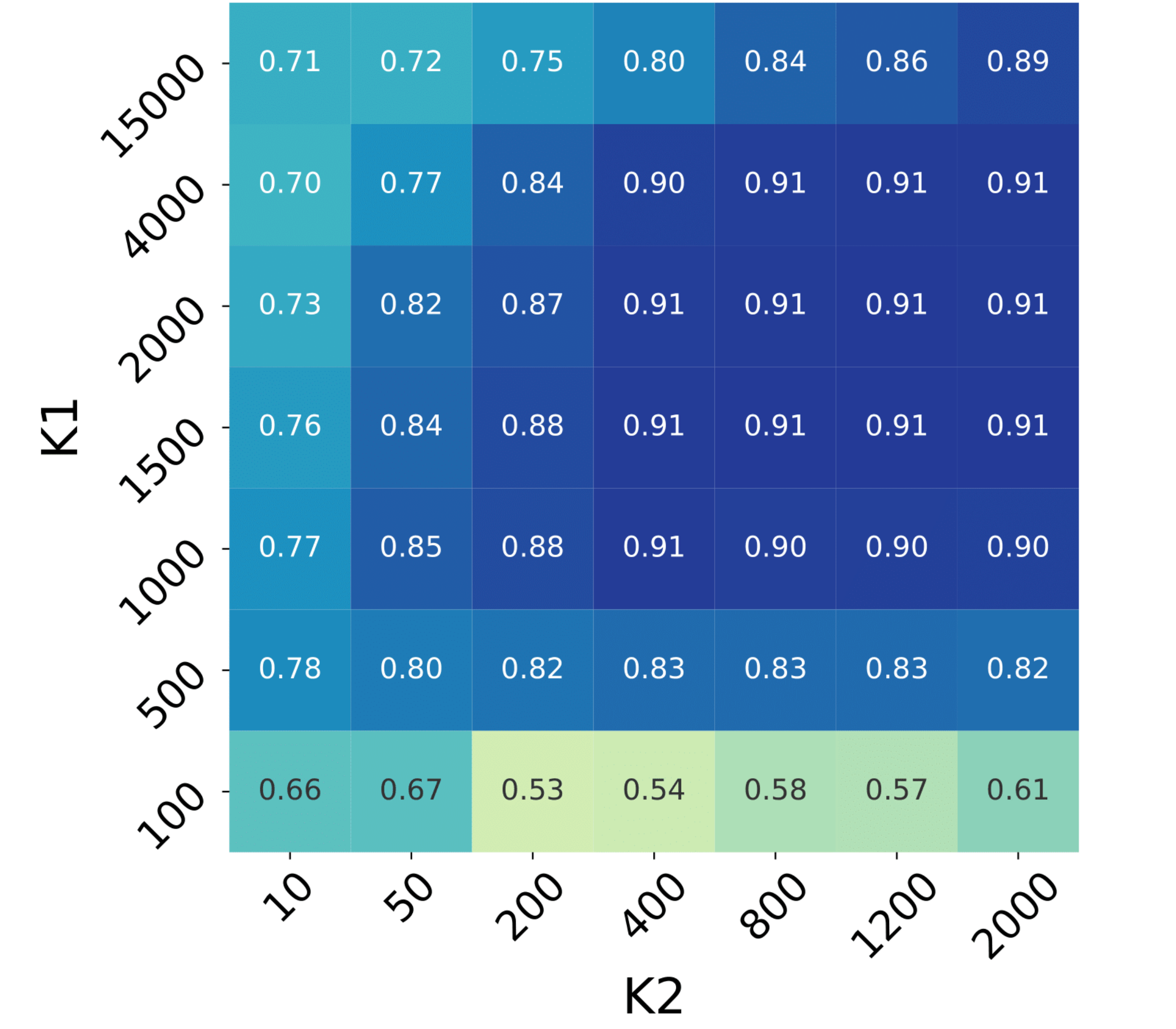}
          \vskip -0.2em
          \caption{\small{Language Score}}
      \end{subfigure}
            \vskip -0.5em
          \caption{Sweep over length of interactive learning phase $k_1$ and length of imitation phase $k_2$ on the Lewis game (darker is higher). Low or high $k_1$ result in poor task and language score. Similarly, low $k_2$ induces poor results while high $k_2$ do not reduce performance as one would expect.}
              \label{fig:lewis_heatmap}
              \vskip -1em
      \end{figure}
\begin{figure}[t]  
          \begin{subfigure}[b]{0.5\columnwidth}
          \centering
          \includegraphics[width=1\linewidth]{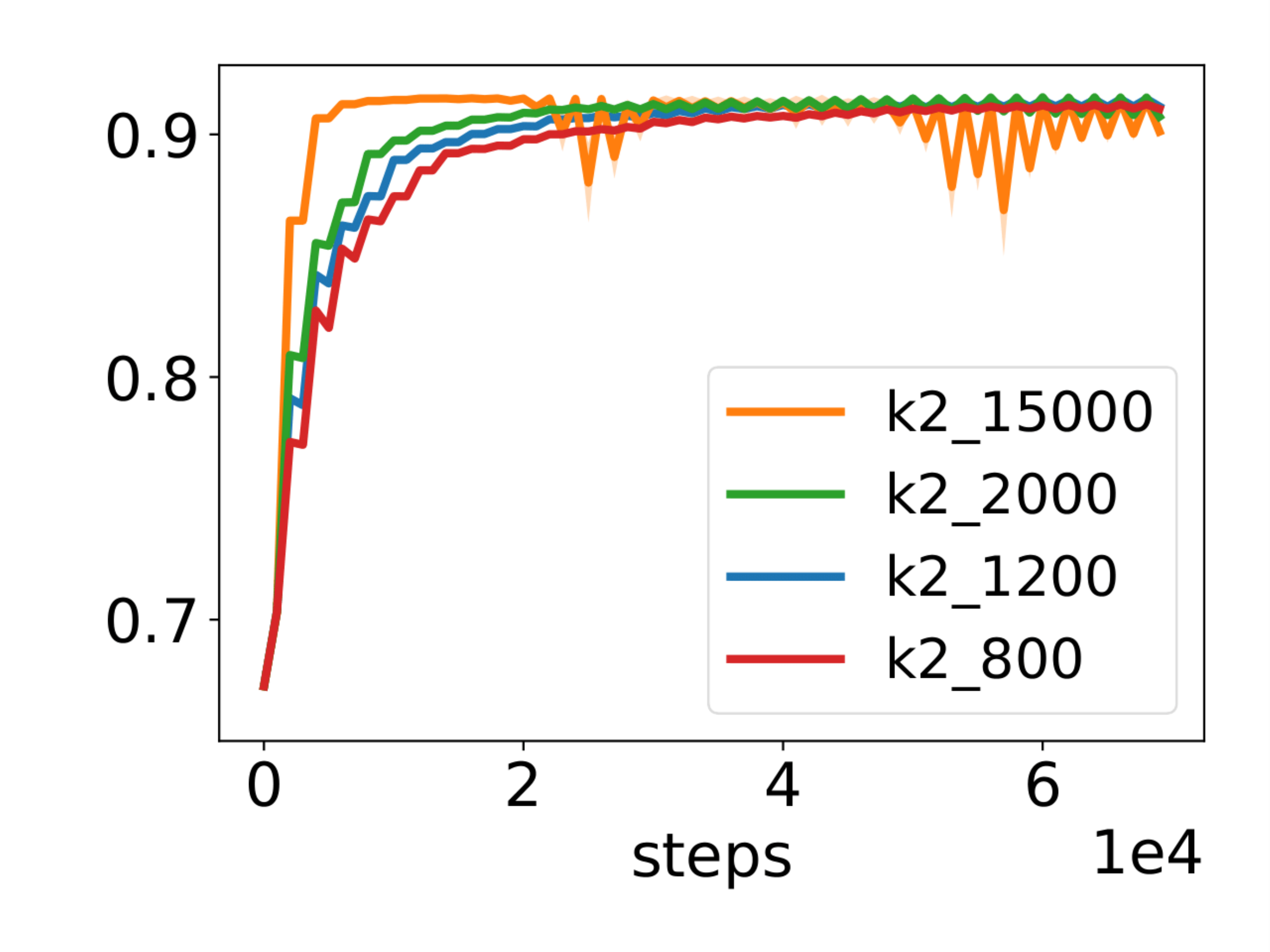}
          \vskip -0.5em
          \caption{argmax}
          \end{subfigure}
         \hspace{-0.5em}%
        \begin{subfigure}[b]{0.5\columnwidth}
          \centering
          \includegraphics[width=1\linewidth]{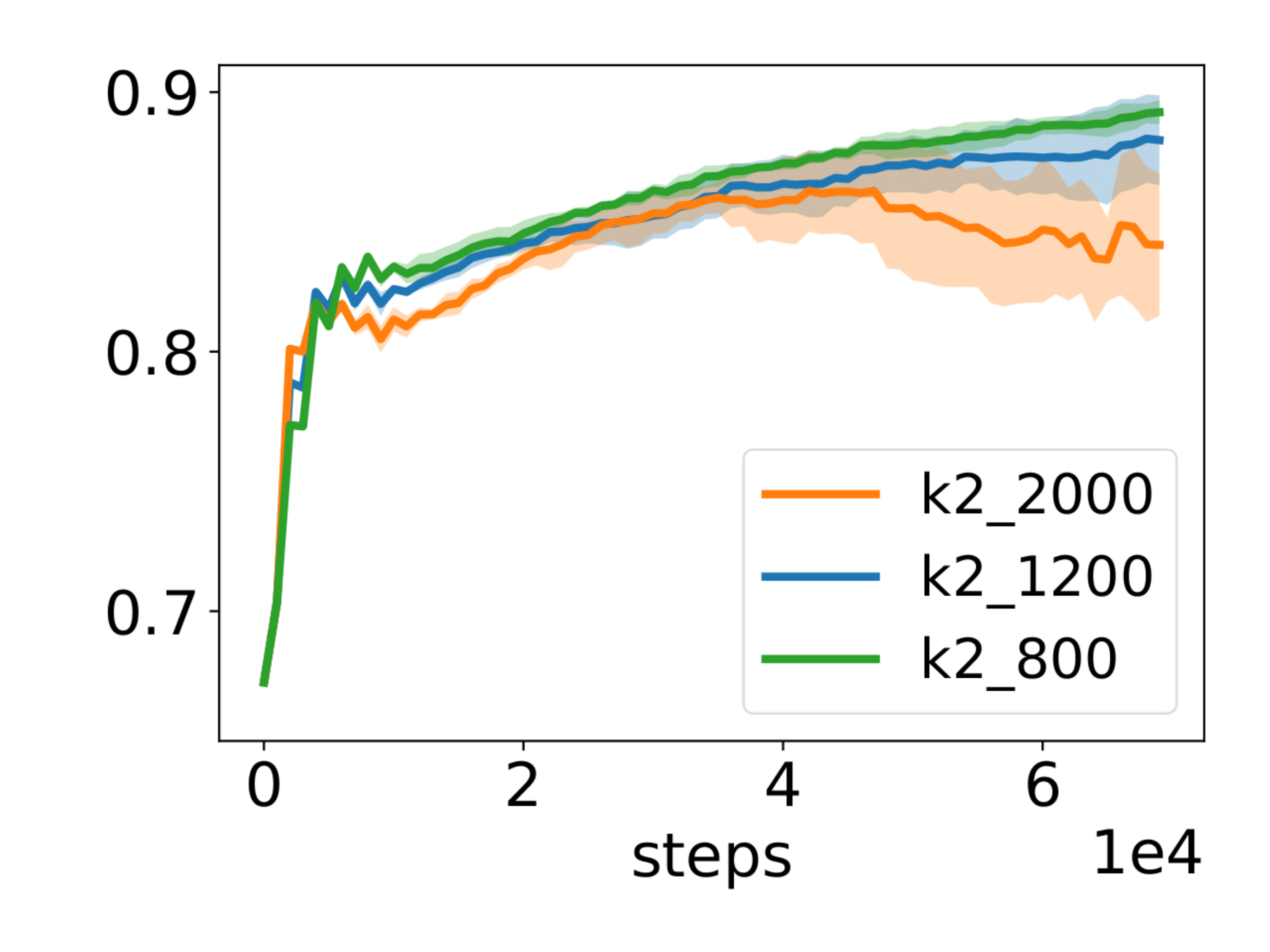}
          \vskip -0.5em
          \caption{KL Minimization}
        \end{subfigure}
                  \vskip -0.5em
    \caption{Language score for different $k_2$ by imitating greedy sampling with cross-entropy (Left) vs distilling the teacher distribution with KL minimization (Right). As distillation relaxes the learning bottleneck, we observe a drop in language score with overfitting when the student imitation learning length increases.}
                      \vskip -0.5em
    \label{fig:k2_evol}
\end{figure}

 
     \paragraph{\SIL Properties} We perform a hyper-parameter sweep for the Lewis Game in Figure~\ref{fig:lewis_heatmap} over the core \SIL parameters, $k_1$ and $k_2$, which are, respectively, the length of interactive and imitation training phase. 
     We simply set $k'_2=k_2$ since in a toy setting the receiver can always  adjust to the sender quickly.
     We find that for each $k_2$, the best $k_1$ is in the middle. This is expected since a small $k_1$ would let the imitation phase constantly disrupt the normal interactive learning, while a large $k_1$ would entail an already drifted teacher.
     We see that $k_2$ must be high enough to successfully transfer teacher distributions to the student. However, when a extremely large $k_2$ is set, we do not observe the expected performance drop predicted by the learning bottleneck: The overfitting of the student to the teacher should reduce \SIL's resistance to language drift. To resolve this dilemma, we slightly modify our imitation learning process. Instead of doing supervised learning on the samples from teachers, we explicitly let student imitate the complete teacher distribution by minimizing $KL(s(;\theta^T) || s(;\theta^S))$. The result is in Figure~\ref{fig:k2_evol}, and we can see that increasing $k_2$ now leads to a loss of performance, which confirms our hypotheses. 
     In conclusion, \SIL has good performance in a (large) valley of parameters, and a proper imitation learning process is also crucial for constructing the learning bottleneck. 

\section{Experiments: The Translation Game}    

    Although being insightful, the Lewis game is missing some core language properties, e.g., word ambiguity or unrealistic word distribution etc. 
    As it relies on a basic finite language, it would be premature to draw too many conclusions from this simple setting~\citep{hayes1988second}. In this section, we present a larger scale application of \SIL in a natural language setting by exploring the translation game~\citep{lee2019countering}.

\begin{figure*}[t]
    \begin{subfigure}[b]{0.25\linewidth}
         \centering
         \includegraphics[width=1.05\linewidth]{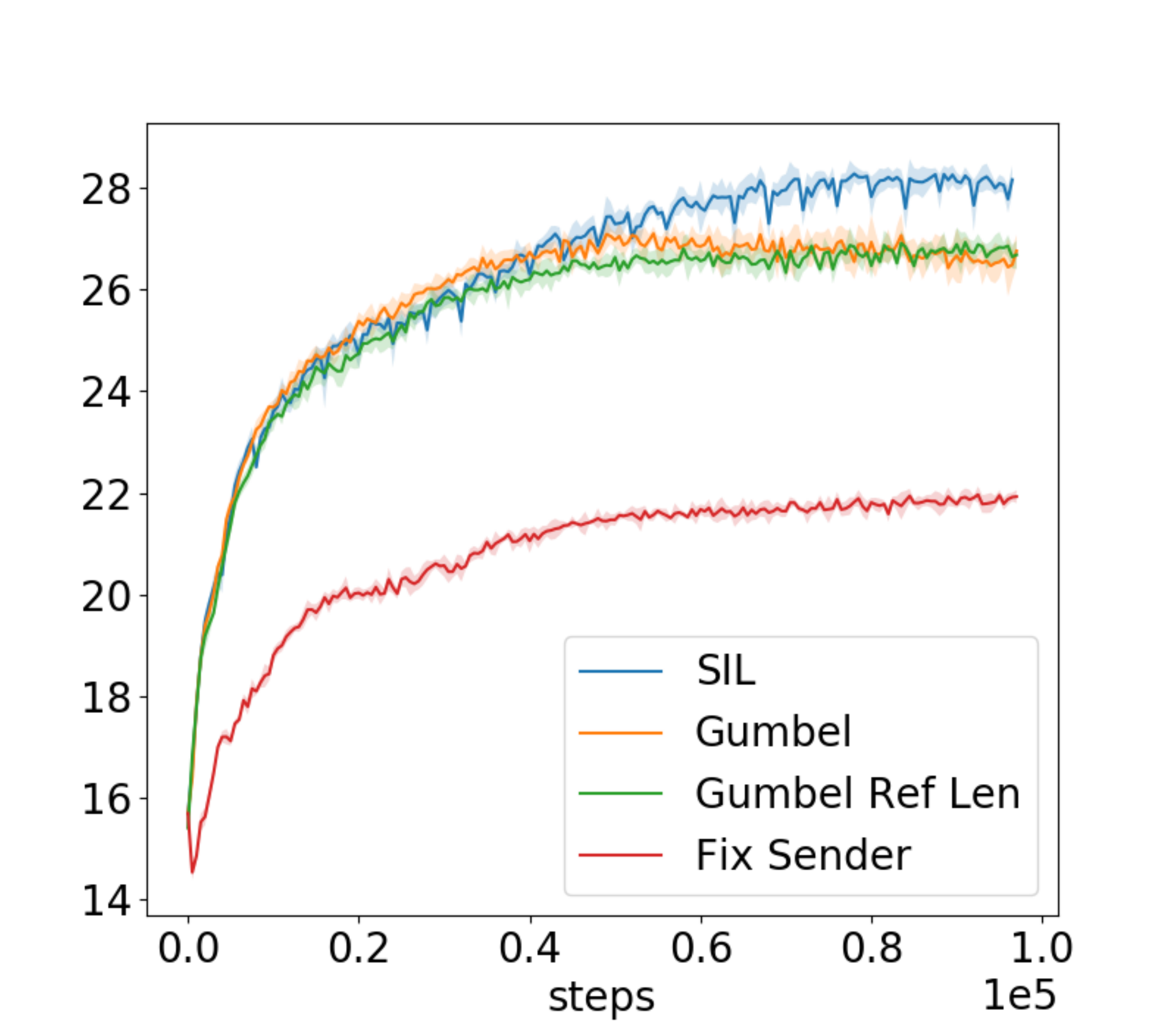}
         \caption{BLEU De (Task Score)}
     \end{subfigure}
     \hspace{-0.5em}%
     \begin{subfigure}[b]{0.25\linewidth}
         \centering
         \includegraphics[width=1.05\linewidth]{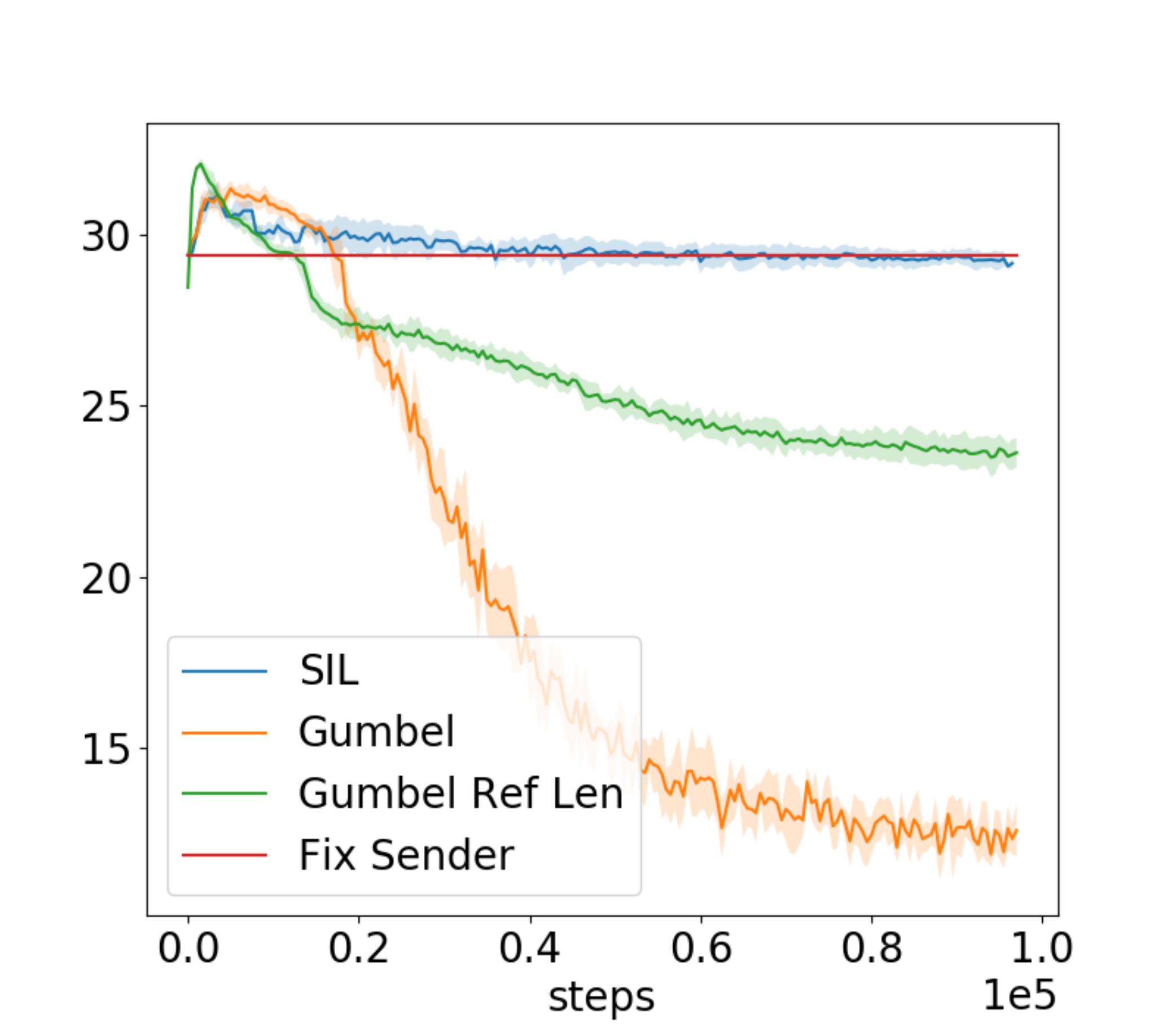}
         \caption{BLEU En}
     \end{subfigure}
             \hspace{-0.5em}%
     \begin{subfigure}[b]{0.25\linewidth}
         \centering
         \includegraphics[width=1.05\linewidth]{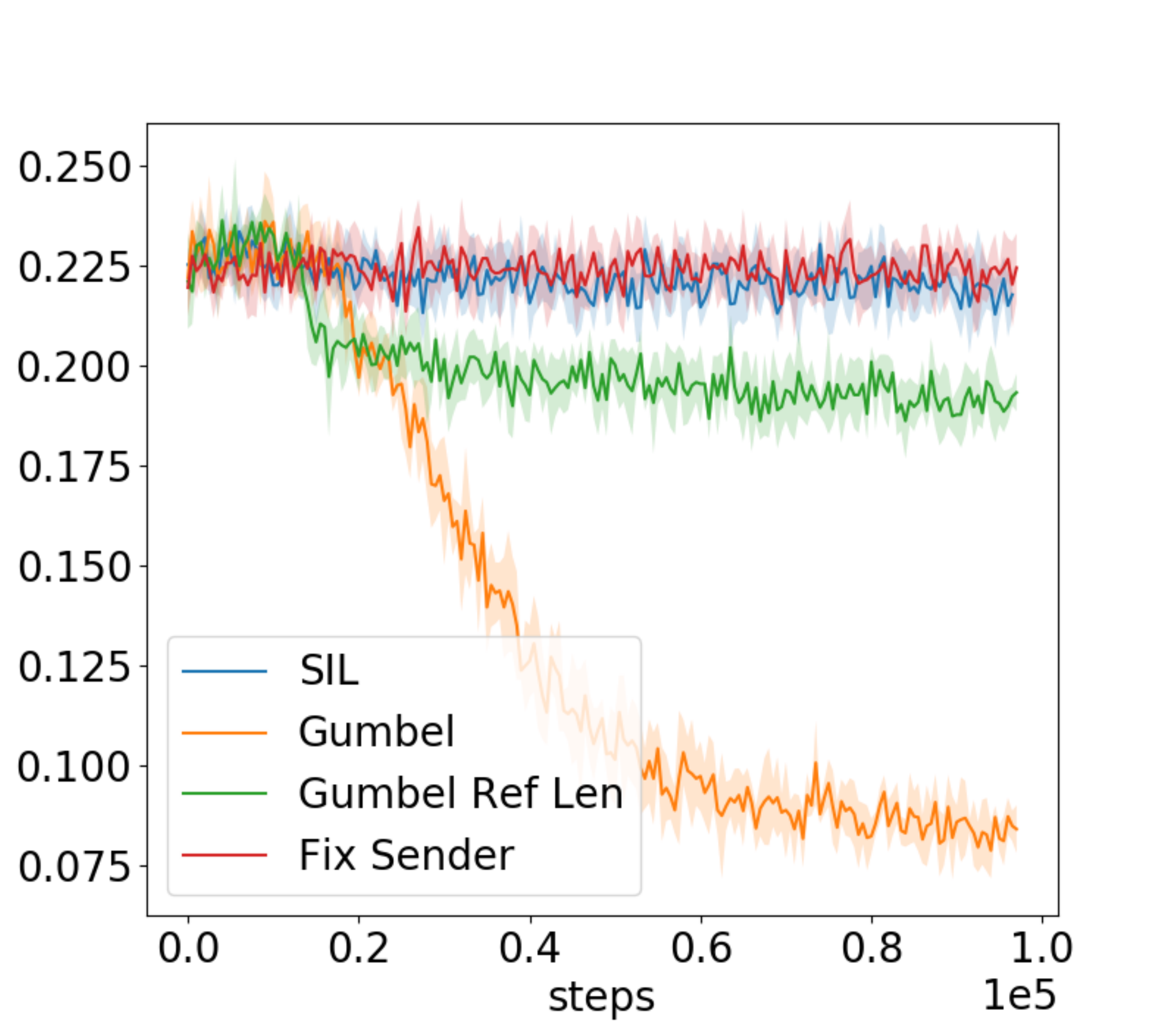}
         \caption{R1}
     \end{subfigure}
        \hspace{-0.5em}%
     \begin{subfigure}[b]{0.25\linewidth}
         \centering
         \includegraphics[width=1.05\linewidth]{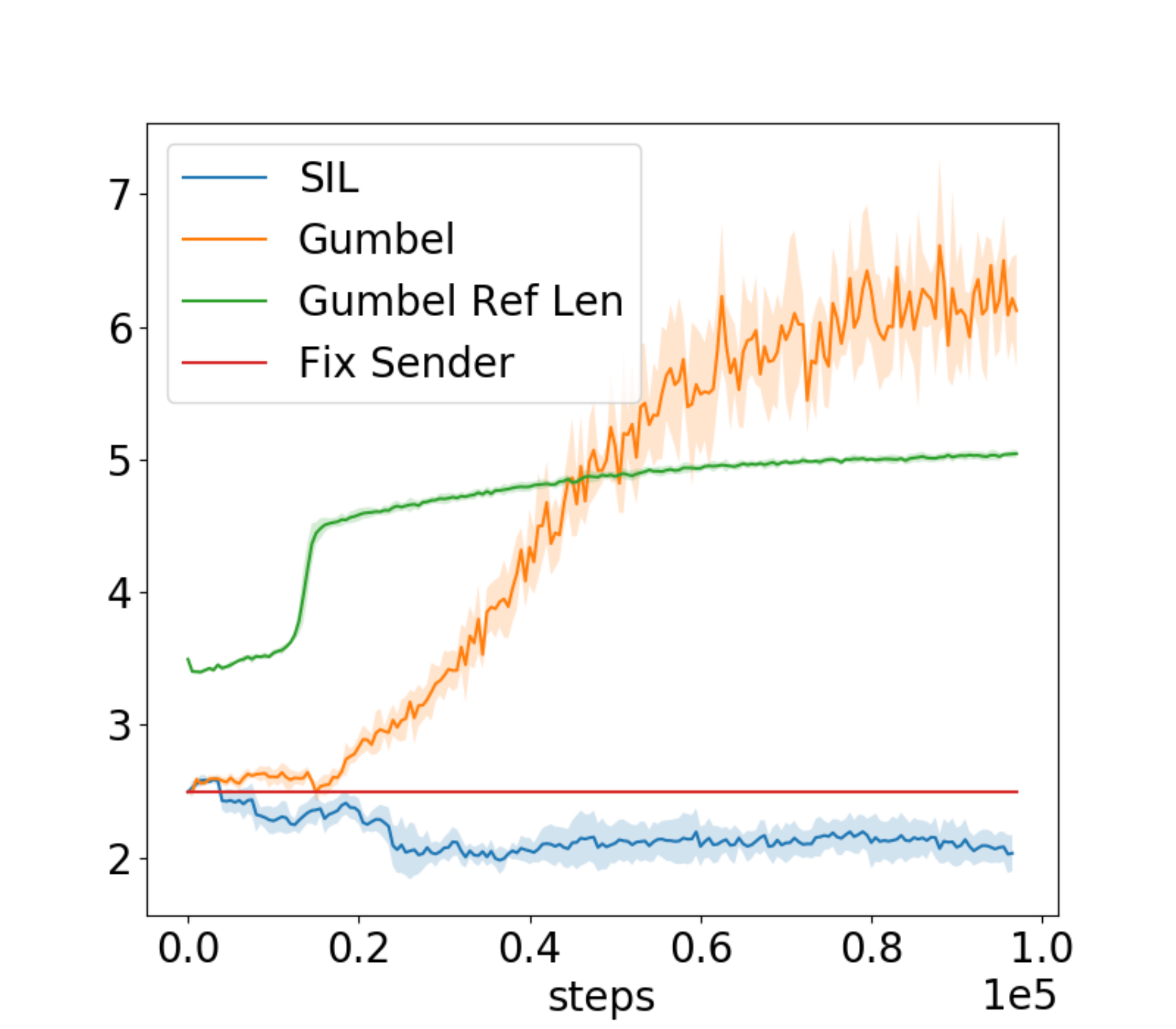}
         \caption{NLL}
     \end{subfigure}
        \hspace{-1em}%
         \caption{The task score and the language score of NIL, S2P, and Gumbel baselines. Fix Sender indicates the maximum performance the sender may achieve without agent co-adaptation. We observe that Gumbel language start drifting when the task score increase. Gumbel Ref Len artificially limits the English message length, which caps the drift. Finally, \SIL manages to both increase language and task score}
    \label{fig:NIL}
    \vskip -1em
\end{figure*}

 \begin{figure}[t]
      \begin{subfigure}[b]{0.5\linewidth}
         \centering
         \includegraphics[width=1.05\linewidth]{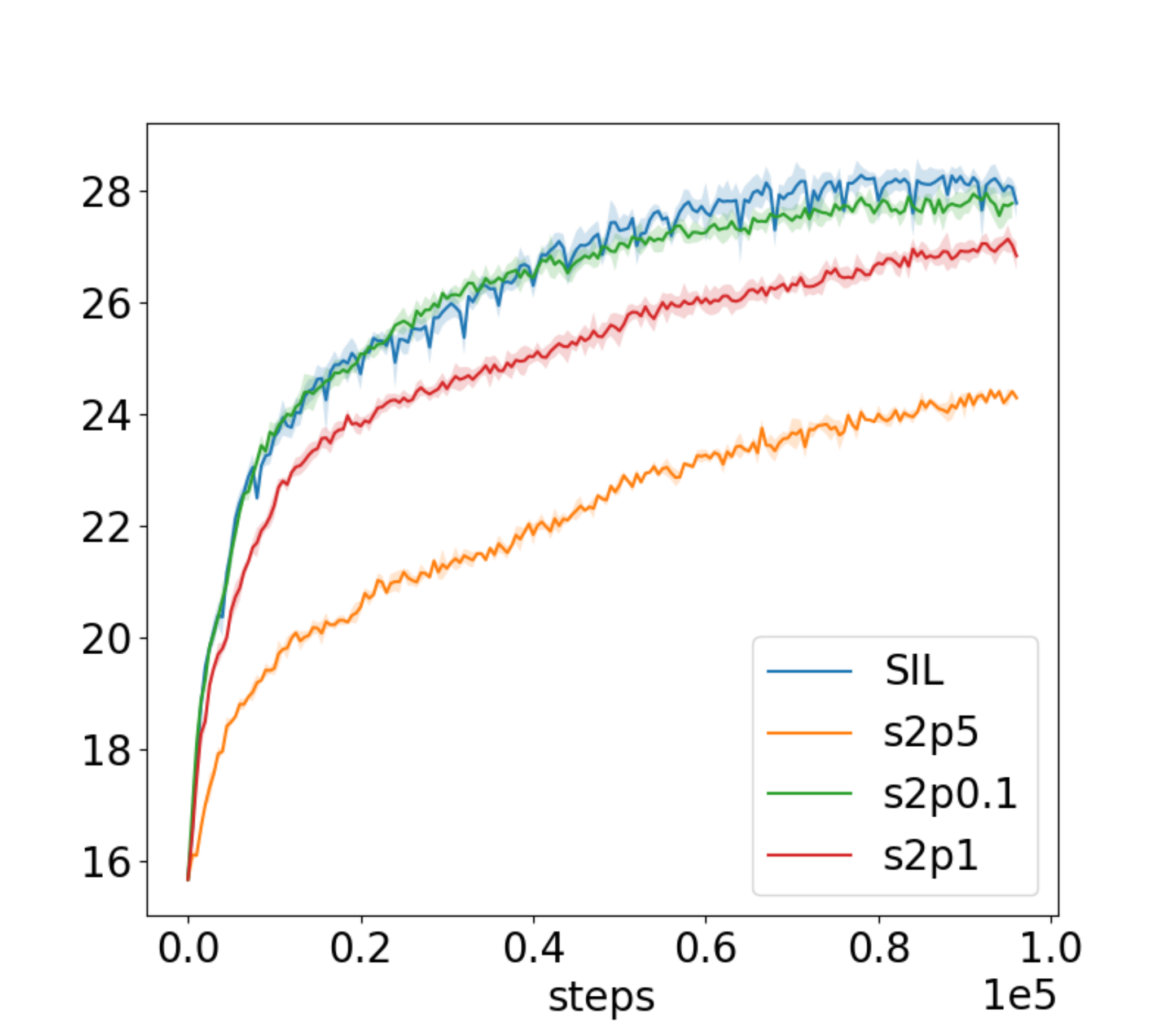}
         \caption{BLEU De (Task Score)}
     \end{subfigure}
             \hspace{-0.5em}
     \begin{subfigure}[b]{0.5\linewidth}
         \centering
         \includegraphics[width=1.05\linewidth]{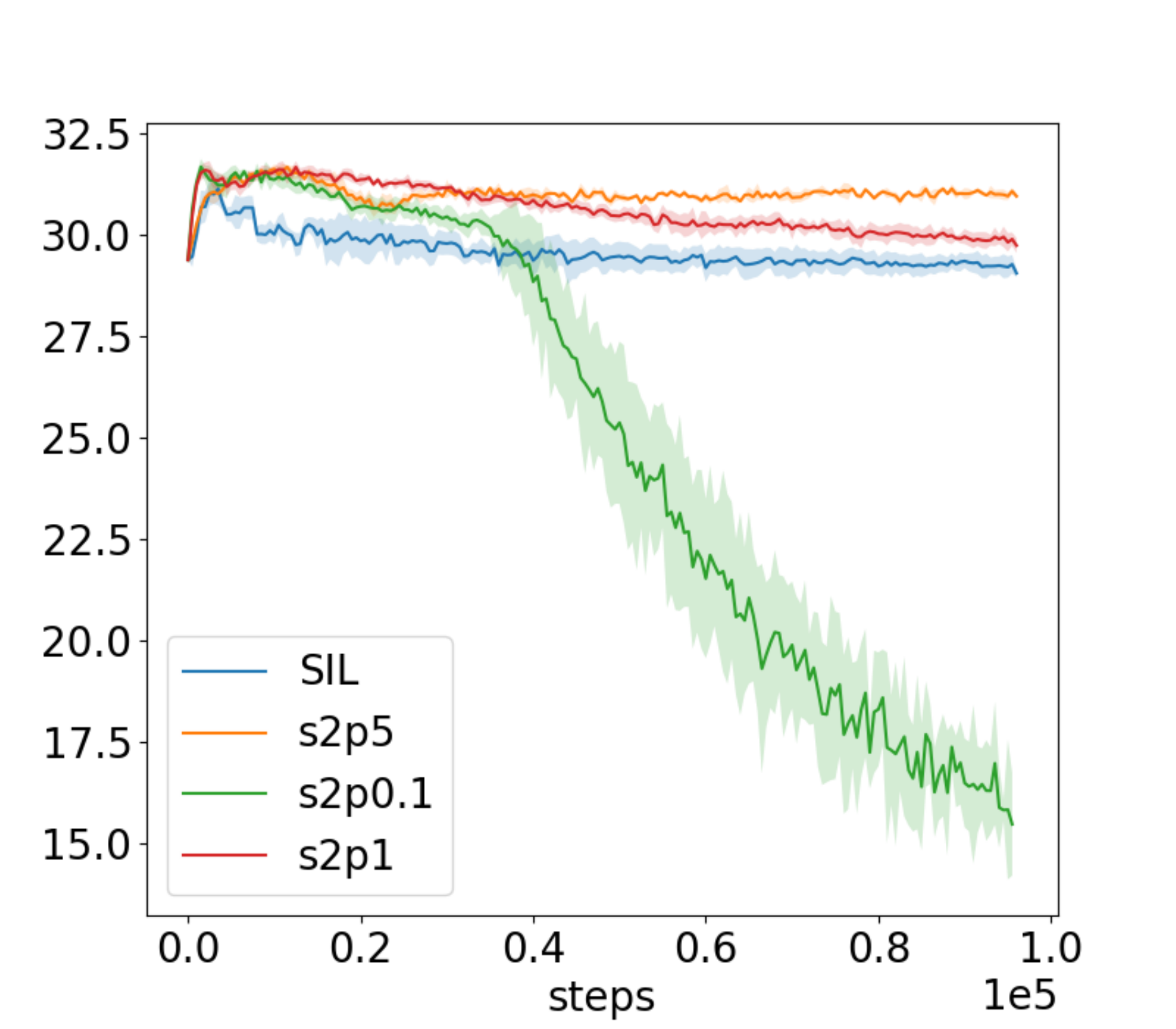}
         \caption{BLEU En}
     \end{subfigure}
     \caption{S2P sweep over imitation loss weight vs. interactive loss. S2P displays a trade-off between a high task score, which requires a low imitation weight, and high language score, which requires high imitation weight. \SIL appears less susceptible to a tradeoff between these metrics}
     \label{fig:s2p_vs_sil}
     \vskip -1em
\end{figure}

    \paragraph{Experimental Setting}\label{sec:translation_setting} The translation game is a S/R game where two agents translate a text from a source language, French (FR), to a target language, German (De), through a pivot language, English (En). This framework allows the evaluation of the English language evolution through translation metrics while optimizing for the Fr$\rightarrow$De translation task, making it a perfect fit for our language drift study.
    
    The translation agents are sequence-to-sequence models with gated recurrent units~\citep{cho2014learning} and attention~\citep{bahdanau2014neural}. First, they are independently pretrained on the IWSLT dataset~\citep{cettolo2012wit3} to learn the initial language distribution.
    The agents are then finetuned with interactive learning by sampling new translation scenarios from the Multi30k dataset~\citep{elliott2016multi30k}, which contains 30k images with the same caption translated in French, English, and German. Generally, we follow the experimental setting of \citet{lee2019countering} for model architecture, dataset, and pre-processing, which we describe in Appendix~\ref{sec:appendix_translation_model_details} for completeness. However, in our experiment, we use GSTE to optimize the sender, whereas \citet{lee2019countering} rely on policy gradient methods to directly maximize the task score. 
    \paragraph{Evaluation metrics} We monitor our task score with \emph{ BLEU(De)}~\citep{papineni2002bleu}, it estimates the quality of the Fr$\rightarrow$De translation by comparing the translated German sentences to the ground truth German. We then measure the sender language score with three metrics. First, we evaluate the overall language drift with the \emph{BLEU(En)} score from the ground truth English captions. As the BLEU score controls the alignment between intermediate English messages and the French input texts, it captures basic syntactic and semantic language variations. Second, we evaluate the structural drift with the negative log-likelihood (\emph{NLL}) of the generated English under a pretrained language model. Third, we evaluate the semantic drift by computing the image retrieval accuracy ($R1$) with a pretrained image ranker; the model fetches the ground truth image given 19 distractors and generated English. The language and image ranker models are further detailed in Appendix~\ref{sec:appendix_translation_lm_ranker}.
    \paragraph{Results} We show our main results in Figure~\ref{fig:NIL}, and a full summary in Table~\ref{table:core} in Appendix~\ref{appendix_translation_game}. Runs are averaged over five seeds and shaded areas are one standard deviation. The x-axis shows the number of interactive learning steps.
    
    After pretraining our language agents on the IWSLT corpus, we obtain the single-agent BLEU score of $29.39$ for Fr$\rightarrow$En and $20.12$ for En$\rightarrow$De on the Multi30k captions. When combining the two agents, the Fr$\rightarrow$De task score drops to $15.7$, showing a compounding error in the translation pipeline. We thus aim to overcome this misalignment between translation agents through interactive learning while preserving an intermediate fluent English language.
    
    As a first step, we freeze the sender to evaluate the maximum task score without agent co-adaptation. The Fix Sender then improves the task score by 5.3 BLEU(De) while artificially maintaining the language score constant. As we latter achieve a higher task score with Gumbel, it shows that merely fixing the sender would greatly hurt the overall task performance.
    
    We observe that the Gumbel agent improves the task score by 11.32 BLEU(De) points but the language score collapse by 10.2 BLEU(En) points, clearly showing language drift while the two agents co-adapt to solve the translation game. \citet{lee2019countering} also constrain the English message length to not exceed the French input caption length, as they observe that language drift often entails long messages. Yet, this strong inductive bias only slows down language drift, and the language score still falls by 6.0 BLEU(En) points. Finally, \SIL improves the task score by 12.6 BLEU(De) while preserving the language score of the pretrained model. Thus, \SIL successfully counters language drift in the translation game while optimizing for task-completion.

    \paragraph{S2P vs \SIL} We compare the S2P and \SIL learning dynamics in Figure~\ref{fig:s2p_vs_sil} and Figure~\ref{fig:s2p_vs_sil_full} in Appendix~\ref{appendix_translation_game}. S2P balances the supervised and interactive losses by setting a weight $\alpha$ for the imitation loss~\citep{lazaridou2016multi}. First, we observe that a low $\alpha$ value, i.e, 0.1,  improves the task score by 11.8 BLEU(De), matching \SIL performances, but the language score diverges. We thus respectively increase $\alpha$ to 1, and 5, which stops the language drift, and even outperforms \SIL language score by 1.2 BLEU(En) points. However, this language stabilization also respectively lowers the task score by 0.9 BLEU(De) and 3.6 BLEU(De) compared to \SIL. In other words, S2P has an inherent trade-off between task score (with low $\alpha$), and language score (with high $\alpha$), whereas \SIL consistently excels on both task and language scores. We assume that S2P is inherently constrained by the initial training dataset.

              \begin{table*}[t]
\centering
\tiny
\begin{tabular}{ l  p{7.5cm}  p{7.5cm}  } 
          &  \quad \emph{\SIL successfully prevent language drift}                            &  \quad \emph{\SIL can remain close to the valid pretrained models }\\
\end{tabular}
\begin{tabular}{ l | p{7.5cm} | p{7.5cm} } 
 Human    & \textbf{two men, one in blue and one in red, compete in a boxing match.}        & \textbf{there are construction workers working hard on a project} \\
 Pretrain & two men, one in blue and the other in red, fight in a headaching game               & there are workers working hard work on a project. \\ 
 Gumbel   & two men one of one in blue and the other in red cfighting in a acacgame......... & there are construction working hard on a project ........... \\ 
 S2P      & two men, one in blue and the other in red, fighting in a kind of a kind.                         & there are workers working hard working on a project .. \\ 
 SIL      & two men, one in blue and the other in red, fighting in a game.                & there are workers working hard on a project . \\
\end{tabular}
\\
\vspace{0.5em}
\begin{tabular}{ l  p{7.5cm}  p{7.5cm}  } 
          &  \quad \emph{\SIL partially recovers the sentence without drifting}                            &  \quad \emph{\SIL/S2P still drift when facing rare word occurrences (shaped lollipop)}\\
\end{tabular}
\begin{tabular}{ l | p{7.5cm} | p{7.5cm} }
 Human    & \textbf{a group of friends lay sprawled out on the floor enjoying their time together.}     & \textbf{a closeup of a child's face eating a blue , heart shaped lollipop.}\\ 
 Pretrain & a group of friends on the floor of fun together.                                            & a big one 's face plan a blue box.\\ 
 Gumbel   & a group of defriends comadeof on the floor together of of of of of together...............  & a big face of a child eating a blue th-acof of of of chearts....... \\ 
 S2P      & a group of friends of their commodities on the floor of fun together.                       & a big face plan of eating a blue of the kind of hearts.\\ 
 SIL      & a group of friends that are going on the floor together.                                    & a big plan of a child eating a blue datadof the datadof the datadof the data@@ \\

\end{tabular}
\caption{Selected generated English captions. Vanilla  Gumbel drifts by losing grammatical structure, repeating patches of words, and inject noisy words. Both S2P and SIL counter language drift by generating approximately correct and understandable sentences. However, they become unstable when dealing with rare word occurrences.}
\label{tab:samples}
\vskip -0em
\end{table*} 

    \paragraph{Syntactic and Semantic Drifts} As described in Section~\ref{sec:translation_setting}, we attempt to decompose the Language Drift into syntactic drifts, by computing language likelihood ($NLL$), and semantic drifts, by aligning images and generated captions ($R1$). In Figure~\ref{fig:NIL}, we observe a clear correlation between those two metrics and a drop in the language BLEU(En) score. For instance, Vanilla-Gumbel simultaneously diverges on these three scores, while the sequence length constraint caps the drifts. We observe that \SIL does not improve language semantics, i.e., $R1$ remains constant during training, whereas it produces more likely sentences as the $NLL$ is improved by 11\%. Yet, S2P preserves slightly better semantic drift, but its language likelihood does not improve as the agent stays close to the initial distribution.

\begin{figure}[t]
    \centering
  \includegraphics[width=0.9\linewidth]{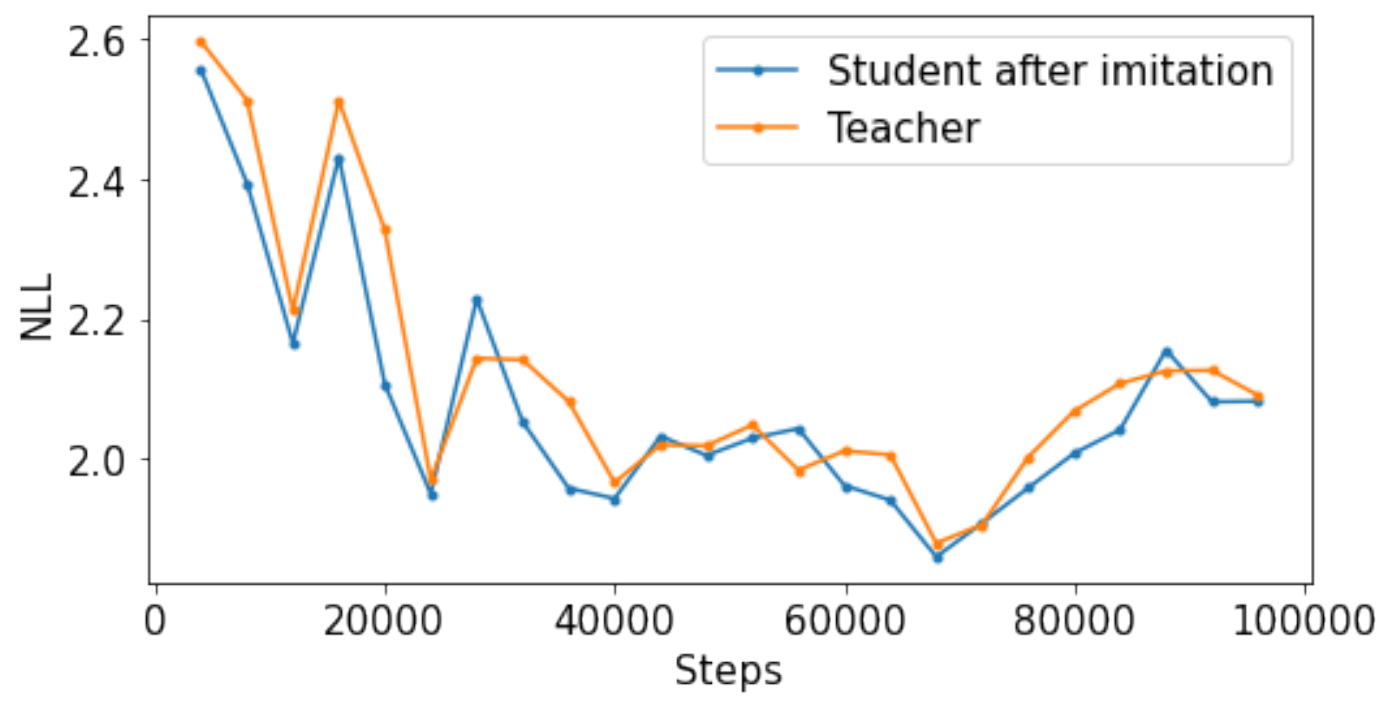}
  \caption{$NLL$ of the teacher and the student after imitation learning phase. In the majority of iterations, the student after imitation obtains a lower NLL than the teacher, after supervised 
  training on the teacher's generated data.}
  \label{fig:lm_score_desc}
  \vskip -1em
\end{figure}
    
\begin{figure}[t]
\vskip -1em
    \centering
         \begin{subfigure}[b]{0.5\columnwidth}
         \centering
         \includegraphics[width=1.05\linewidth]{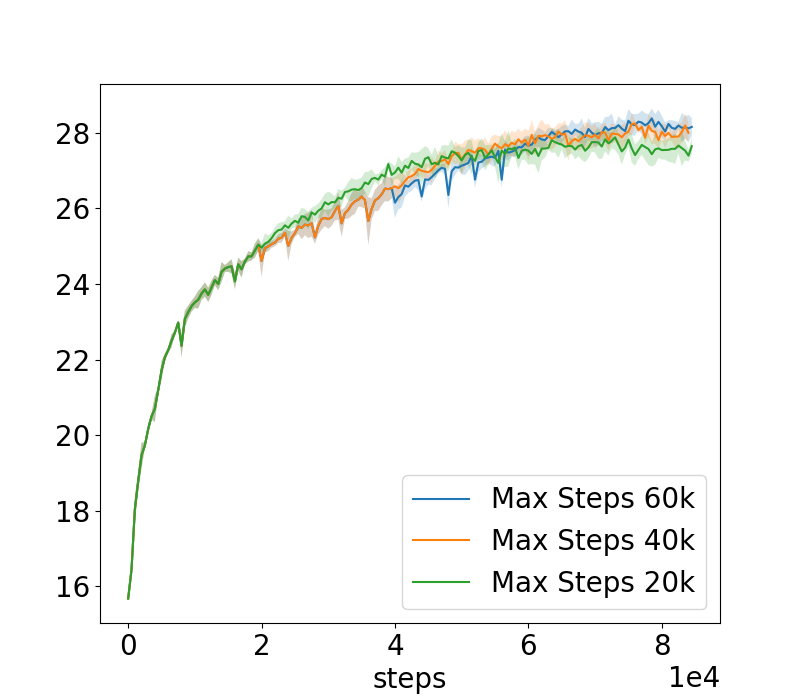}
         \caption{BLEU De}
     \end{subfigure}
    \hspace{-0.5em}%
         \begin{subfigure}[b]{0.5\columnwidth}
         \centering
         \includegraphics[width=1.05\linewidth]{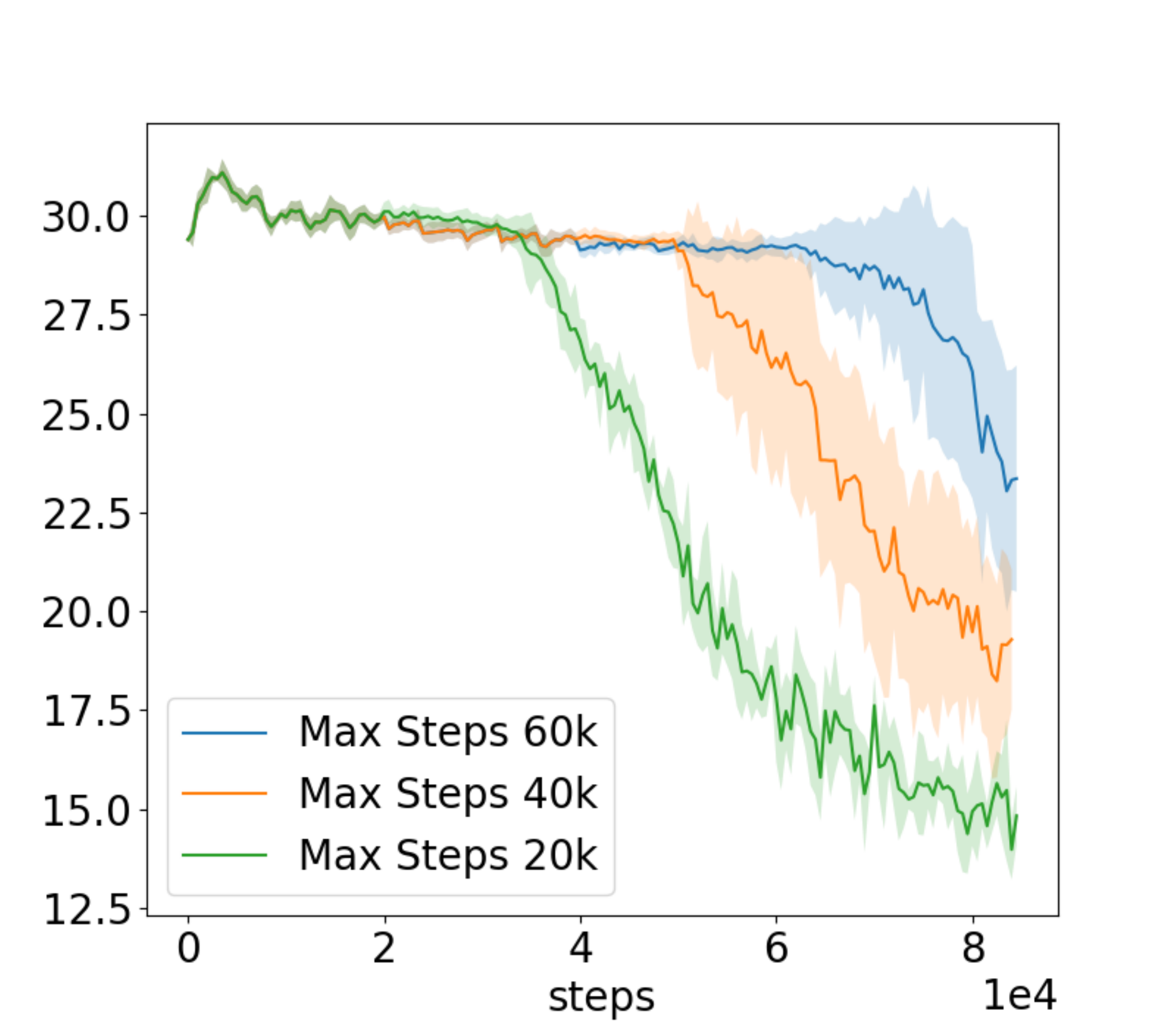}
         \caption{BLEU En}
     \end{subfigure}
         \hspace{-0.5em}%
    \caption{Effect of stopping \SIL earlier in the training process. \SIL maximum steps set at 20k, 40k and 60k. \SIL appears to be important in preventing language drift through-out training.}
    \label{fig:max_steps}
\end{figure}

    \paragraph{\SIL Mechanisms} We here verify the initial motivations behind \SIL by examining the impact of the learning bottleneck in Figure~\ref{fig:lm_score_desc} and the structure-preserving abilities of \SIL in Figure~\ref{fig:max_steps}.
    As motivated in Section~\ref{sec:sil}, each imitation phase in the \SIL aims to filtering-out emergent unstructured language by generating an intermediate dataset to train the student. 
    To verify this hypothesis, we examine the change of negative language likelihood ($NLL$) from the teacher to the student after imitation. We observe that after imitation, the student consistently improves the language likelihood of its teacher, indicating a more regular language production induced by the imitation step. In another experiment, we stop the iterated learning loop after 20k, 40k and 60k steps and continue with standard interactive training. We observe that the agent's language score starts dropping dramatically as soon as we stop \SIL while the task score keep improving. This finding supports the view that \SIL persists in preventing language drift throughout training, and that the language drift phenomenon itself appear to be robust and not a result of some unstable initialization point. 
    \paragraph{Qualitative Analysis}  In Table~\ref{tab:samples}, we show some hand-selected examples of English messages from the translation game. As expected, we observe that the vanilla Gumbel agent diverges from the pretrained language models into unstructured sentences, repeating final dots or words. It also introduce unrecognizable words such as "cfighting" or "acacgame" by randomly pairing up sub-words whenever it faces rare word tokens. S2P and \SIL successfully counter the language drift, producing syntactically valid language. However, they can still produce semantically inconsistent captions, which may be due to the poor pretrained model, and the lack of grounding~\citep{lee2019countering}. 
    Finally, we still observe language drift when dealing with rare word occurrences. 
    Additional global language statistics can be found in Appendix that supports that \SIL preserves language statistical properties.
        
  \section{Conclusion}
In this paper we proposed a method to counter language
drift in task-oriented language settings. The method, named
Seeded Iterated Learning is based on the broader principle
of iterated learning. It alternates imitation learning and
task optimisation steps. We modified the iterated learning
principle so that it starts from a seed model trained on actual
human data, and preserve the language properties during
training. Our extensive experimental study revealed that
this method outperforms standard baselines both in terms
of keeping a syntactic language structure and of solving the
task. As future work, we plan to test this method on complex
dialog tasks involving stronger cooperation between agents.
  
\section*{Acknowledgement}
We thank the authors of the paper \emph{Countering Language Drift via Visual Grounding}, i.e, Jason Lee, Kyunghyun Cho, Douwe Kiela for sharing their original codebase with us. We thank Angeliki Lazaridou for her multiple insightful guidance alongside this project. We also thank  Anna Potapenko, Olivier Tieleman and Philip Paquette for helpful discussions. This research was enabled in part by computations support provided by Compute Canada (\url{www.computecanada.ca}).

{\small
\bibliography{biblio}
\bibliographystyle{icml2020}
\clearpage
} 
\appendix

\onecolumn

\section{Complementary Theoretical Intuition for SIL and Its Limitation}

We here provide a complementary intuition of Seeded Iterated Learning by referring to some mathematical tools that were used to study Iterated Learning dynamics in the general case. These are not the rigorous proof but guide the design of SIL.

One concern is that, since natural language is not fully compositional, whether iterated learning may favor the emergence of a new compositional language on top of the initial one. In this spirit, \citet{griffiths2005bayesian, kalish2007iterated} modeled iterated learning as a Markov Process, and showed that vanilla iterated learning indeed converges to a language distribution that (i) is independent of the initial language distribution, (ii) depends on the student language before the inductive learning step.

Fortunately, \citet{chazelle2016self} show iterated learning can converge towards a distribution close to the initial one with high probability if the intermediate student distributions remain close enough of their teacher distributions and if the number of training observations increases logarithmically with the number of iterations.

This theoretical result motivates one difference between our framework and classical iterated learning: as we want to preserve the pretrained language distribution, we do not initialize the new students from scratch as in~\citep{li2019ease,guo2019emergence,ren2020compositional} because the latter approach exert a uniform prior on the space of language, while we would like to add a prior that favors natural language (e.g. favoring language whose token frequency satisfies Zipf's Law). 

A straightforward instantiation of the above theoretic results is to initialize new students as the pretrained model. However we empirically observe that, periodically resetting the model to initial pretrained model would quickly saturate the task score. As a result, we just keep using the students from the last imitation learning for the beginning of new generation, as well as retain the natural language properties from pretraining checkpoint. 

However, we would also point out the limitation of existing theoretical results in the context of deep learning: 
The theoretical iterated learning results assume the agent to be perfect Bayesian learner (e.g. Learning is infering the posterior distribution of hypothesis given data). However, we only apply standard deep learning training procedure in our setup, which might not have this property.
Because of the assumption of perfect Bayesian learner, \citep{chazelle2019iterated} suggests to use training sessions with increasing length. However in practice, increasing $k_2$ may be counter-productive because of overfitting issues (especially when we have limited number of training scenarios).

\section{Lewis Game}
\label{sec:appendix_lewis}
\subsection{Experiment Details}
\label{sec:appendix_lewis_hyper}
In the Lewis game, the sender and the receiver architecture are 2-layer MLP with a hidden size of 200 and no-activation ($ReLU$ activations lead to similar scores). During interaction learning, we use a learning rate of 1e-4 for \SIL. We use a learning rate of 1e-3 for the baselines as it provides better performance on the language and score tasks. In both cases, we use a training batch size of 100. For the teacher imitation phase, the student uses a learning rate of 1e-4.

In the Lewis game setting, we generate objects with $p=5$ properties, where each property may take $t=5$ values. Thus, it exists 3125 objects, which we split into 3 datasets: the pretraining, the interactive, and testing datasets. The pretraining split only contains 10 combination of objects. As soon as we provide additional objects, the sender and receiver fully solve the game by using the target language, which is not suitable to study the language drift phenomenon. The interactive split contains 30 objects. This choice is arbitrary, and choosing a additional objects gives similar results. Finally, the 3.1k remaining objects are held-out for evaluation. 
 
\subsection{Additional Plots}

\begin{figure}[ht]
    \centering
        \begin{subfigure}[b]{0.32\columnwidth}
          \centering
          \includegraphics[width=1\linewidth]{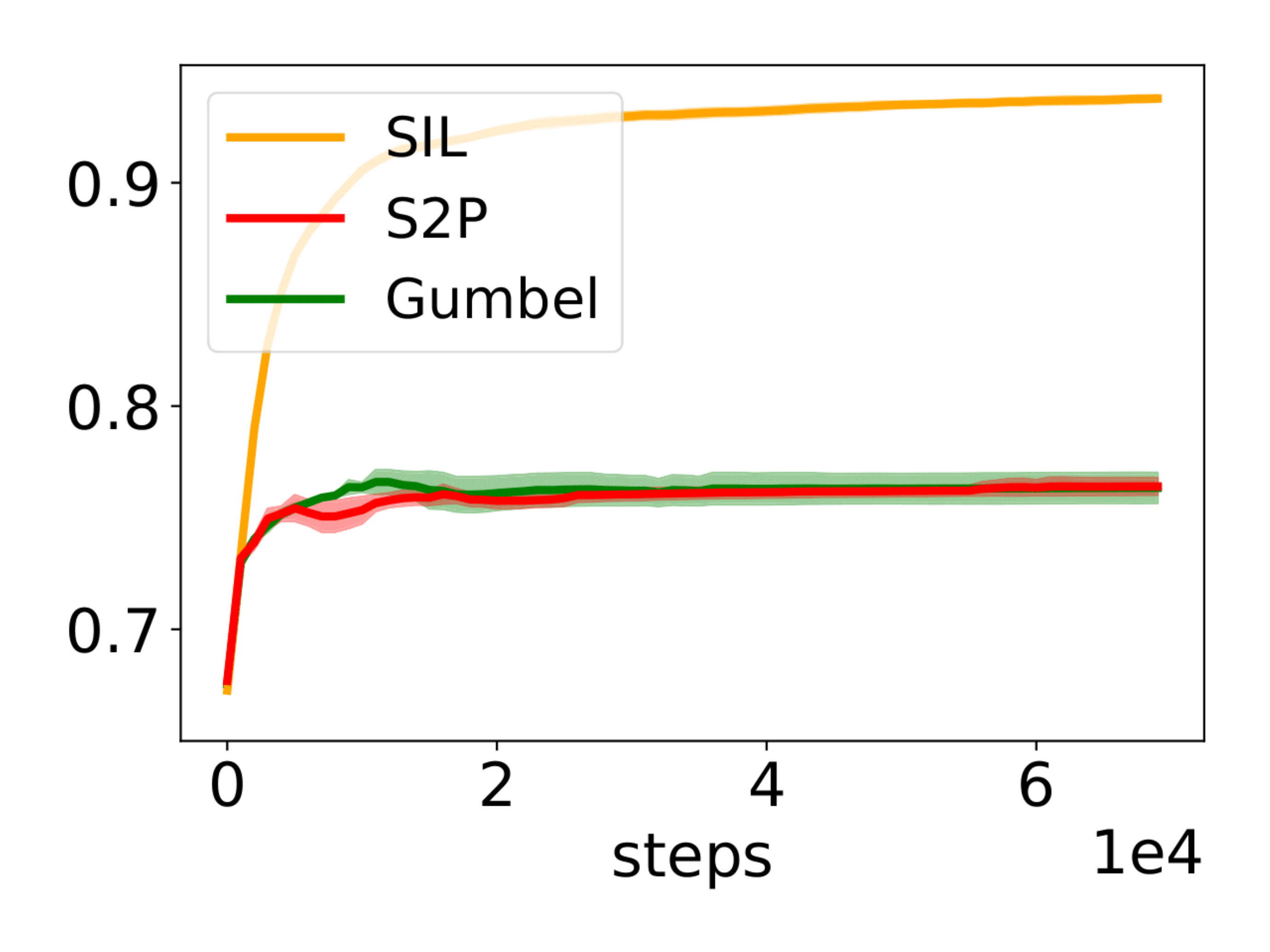}  
          \vskip -0.5em
          \caption{Task Score (Held-Out)}
        \end{subfigure}
        \hfill
        \begin{subfigure}[b]{0.32\columnwidth}
          \centering
          \includegraphics[width=1\linewidth]{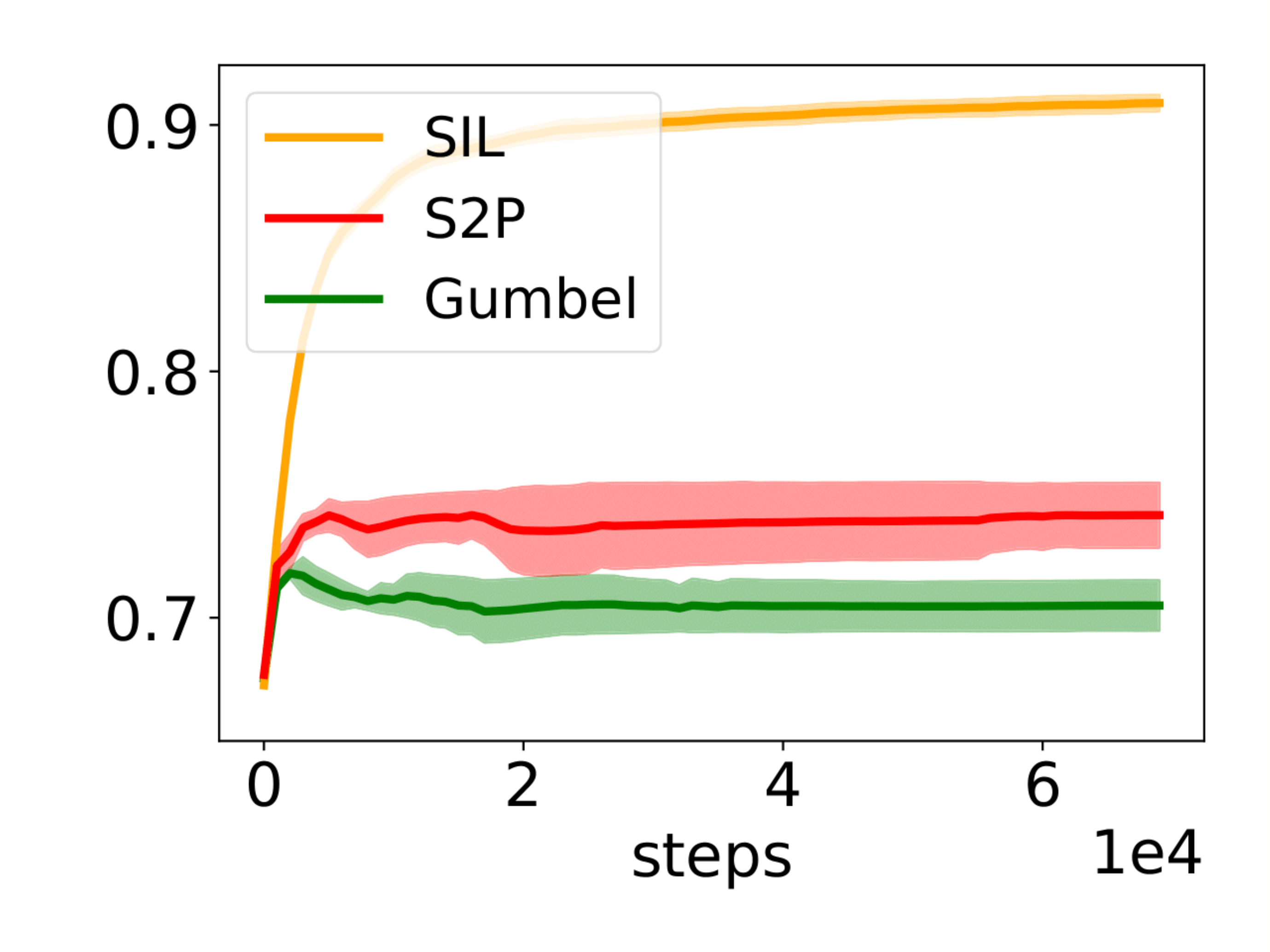}
          \vskip -0.5em
          \caption{Sender Language Score (Held-Out)}
        \end{subfigure}
        \hfill
        \begin{subfigure}[b]{0.32\columnwidth}
          \centering
          \includegraphics[width=\linewidth]{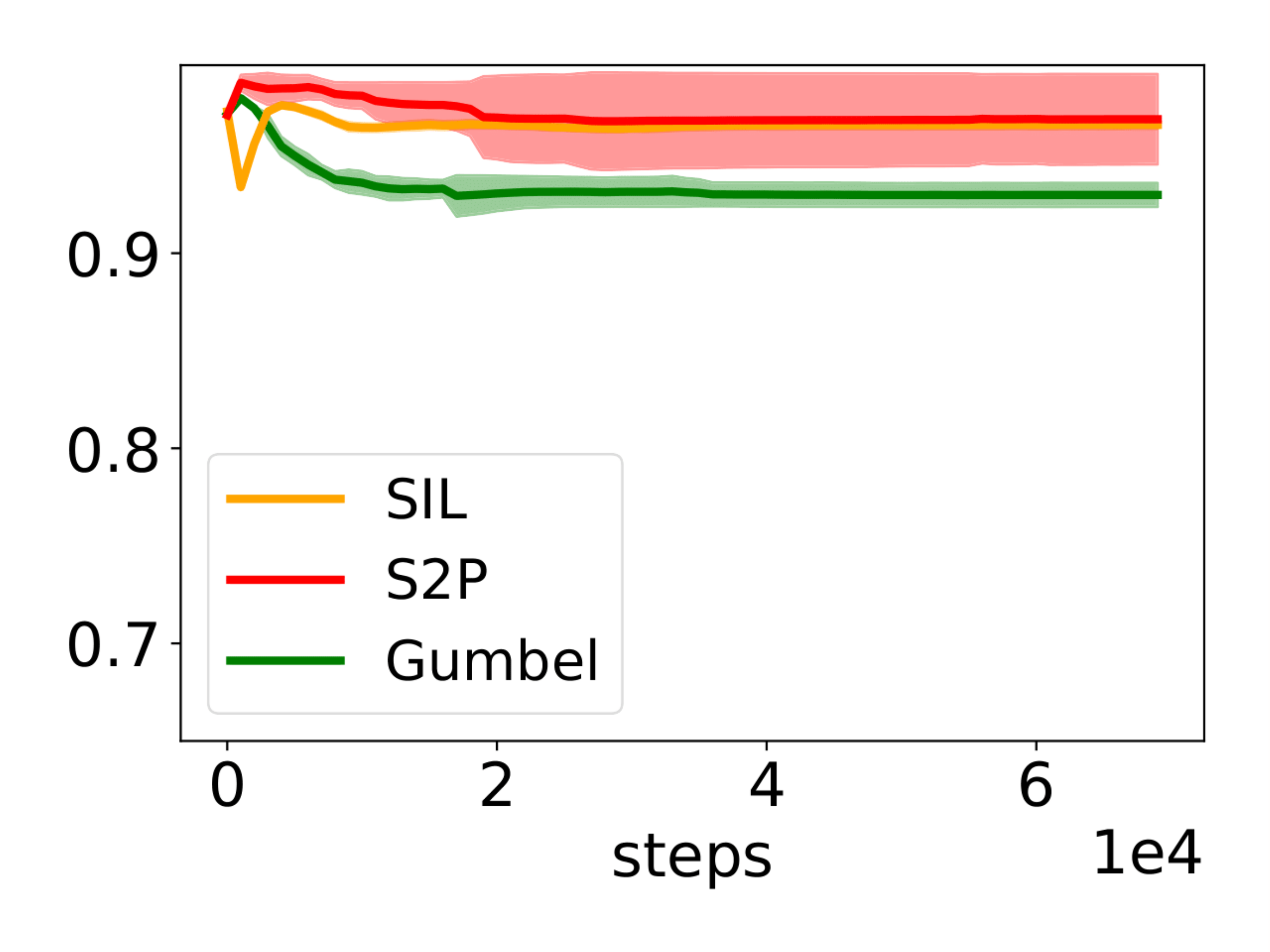}  
          \vskip -0.5em
          \caption{Receiver Language Score (Held-Out)}
        \end{subfigure}
        \\
        \begin{subfigure}[b]{0.32\columnwidth}
          \centering
          \includegraphics[width=\linewidth]{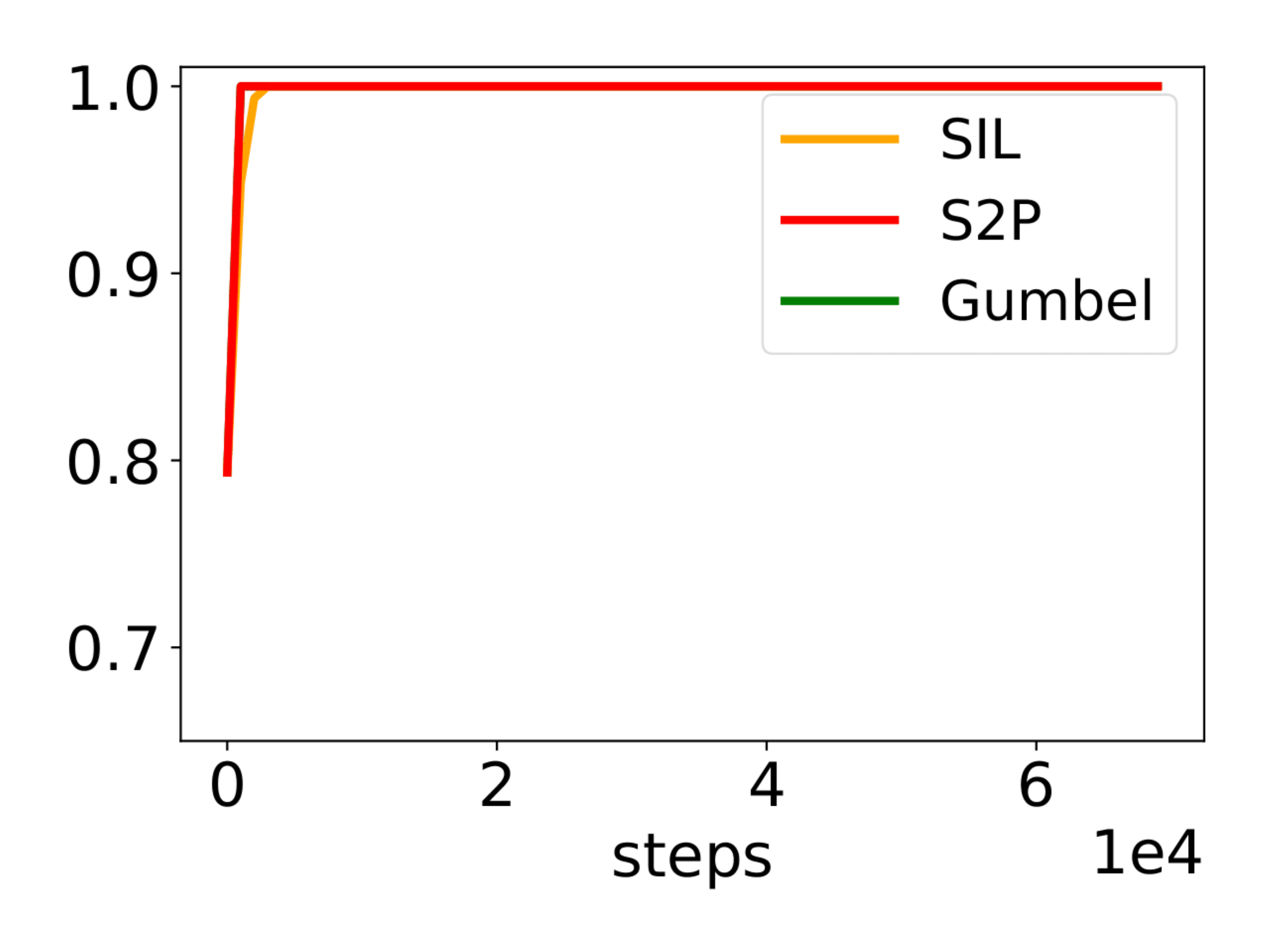}  
          \vskip -0.5em
          \caption{Task Score (Train)}
        \end{subfigure}
        \hfill
        \begin{subfigure}[b]{0.32\columnwidth}
          \centering
          \includegraphics[width=\linewidth]{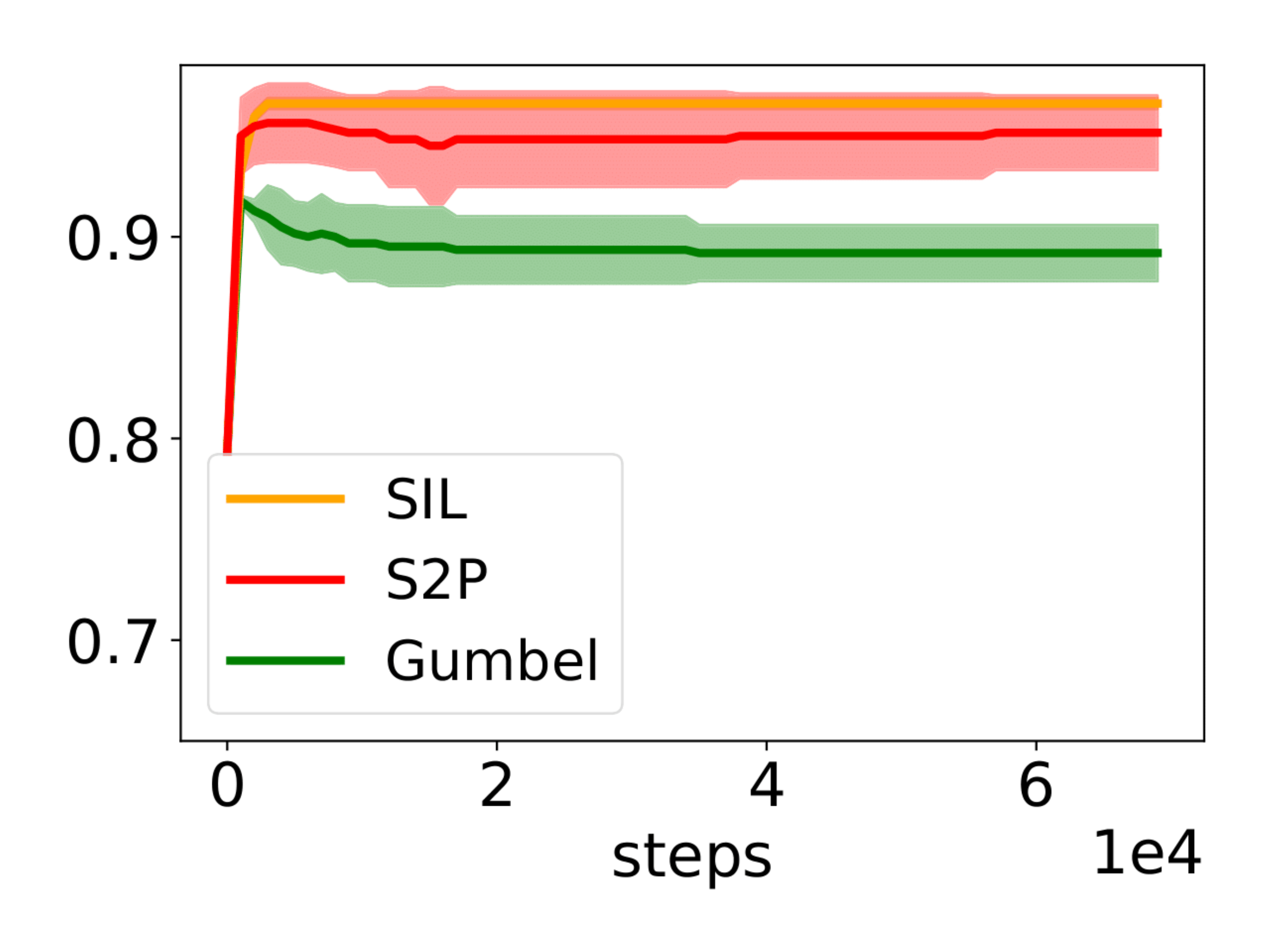}  
          \vskip -0.5em
          \caption{Sender Language Score (Train)}
        \end{subfigure}
        \hfill
        \begin{subfigure}[b]{0.32\columnwidth}
          \centering
          \includegraphics[width=\linewidth]{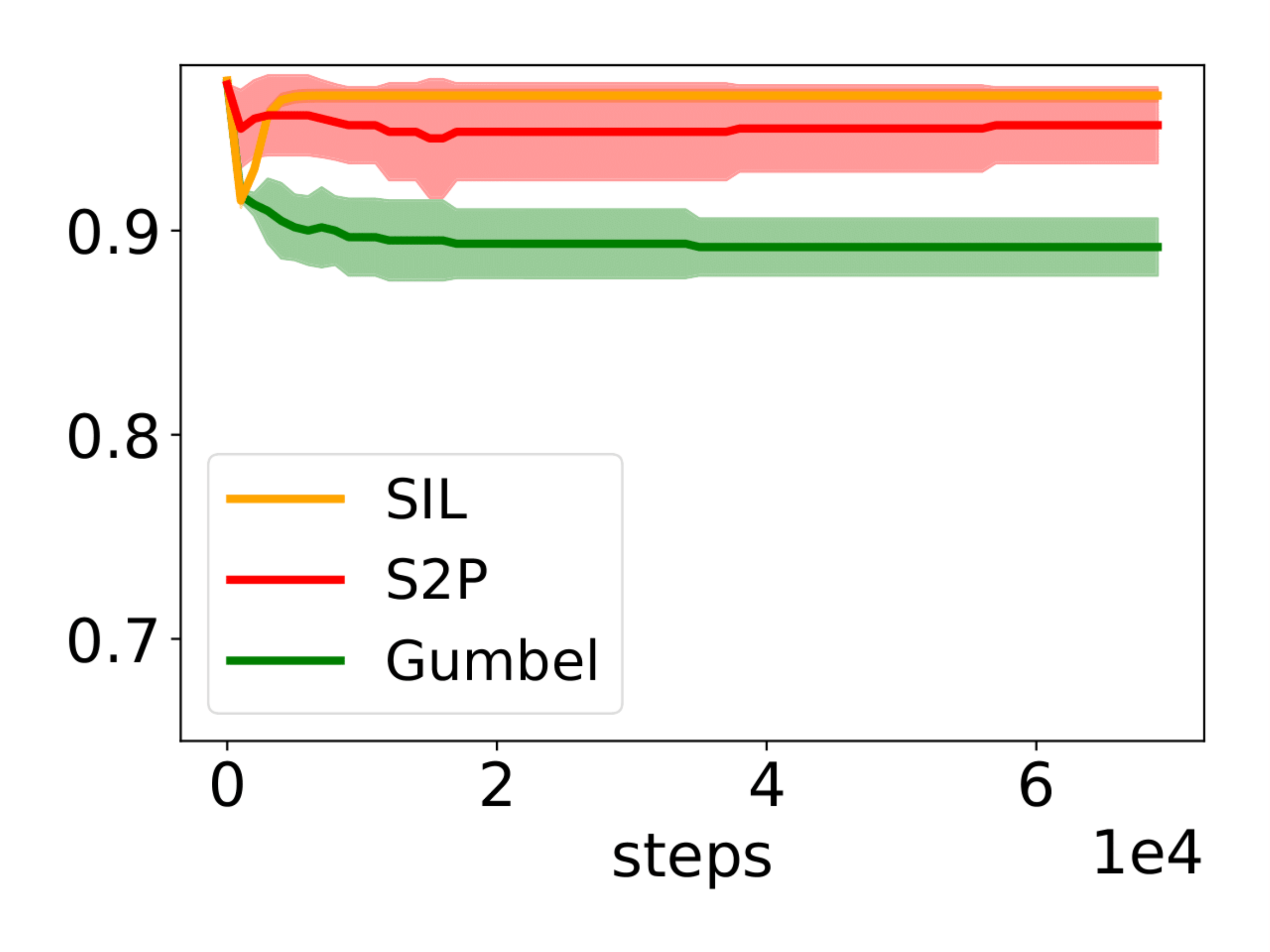} 
          \vskip -0.5em
          \caption{Receiver Language Score (Train)}
        \end{subfigure}
        \caption{Complete training curves for Task score and sender grounding in Lewis Game comparing \SIL vs baselines for $\tau=10$ on the held-out dataset (bottom), and the interactive training split (bottom). We observe that the three methods reach 100\% accuracy on the training task score, but their score differs on the held-out split. For \SIL we use $k_1=1000, k_2=k'_2=400$.}
        \label{fig:full_lewis_curve_10}
\end{figure}

\begin{figure}[h]
    \centering
       \begin{subfigure}[b]{0.32\columnwidth}
          \centering
          \includegraphics[width=1\linewidth]{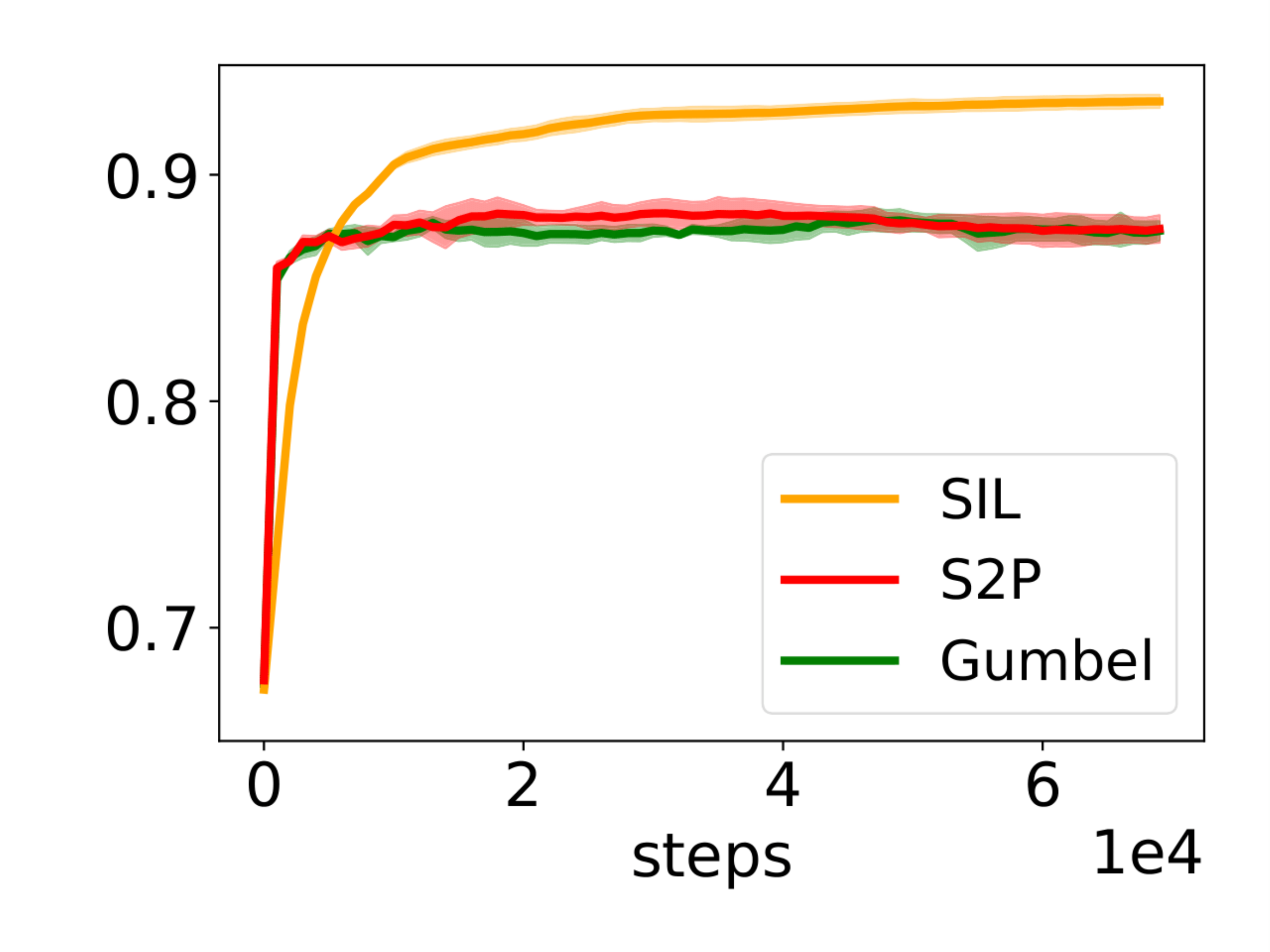} 
          \vskip -0.5em
          \caption{Task Score (Held-Out)}
        \end{subfigure}
        \hfill
        \begin{subfigure}[b]{0.32\columnwidth}
          \centering
          \includegraphics[width=1\linewidth]{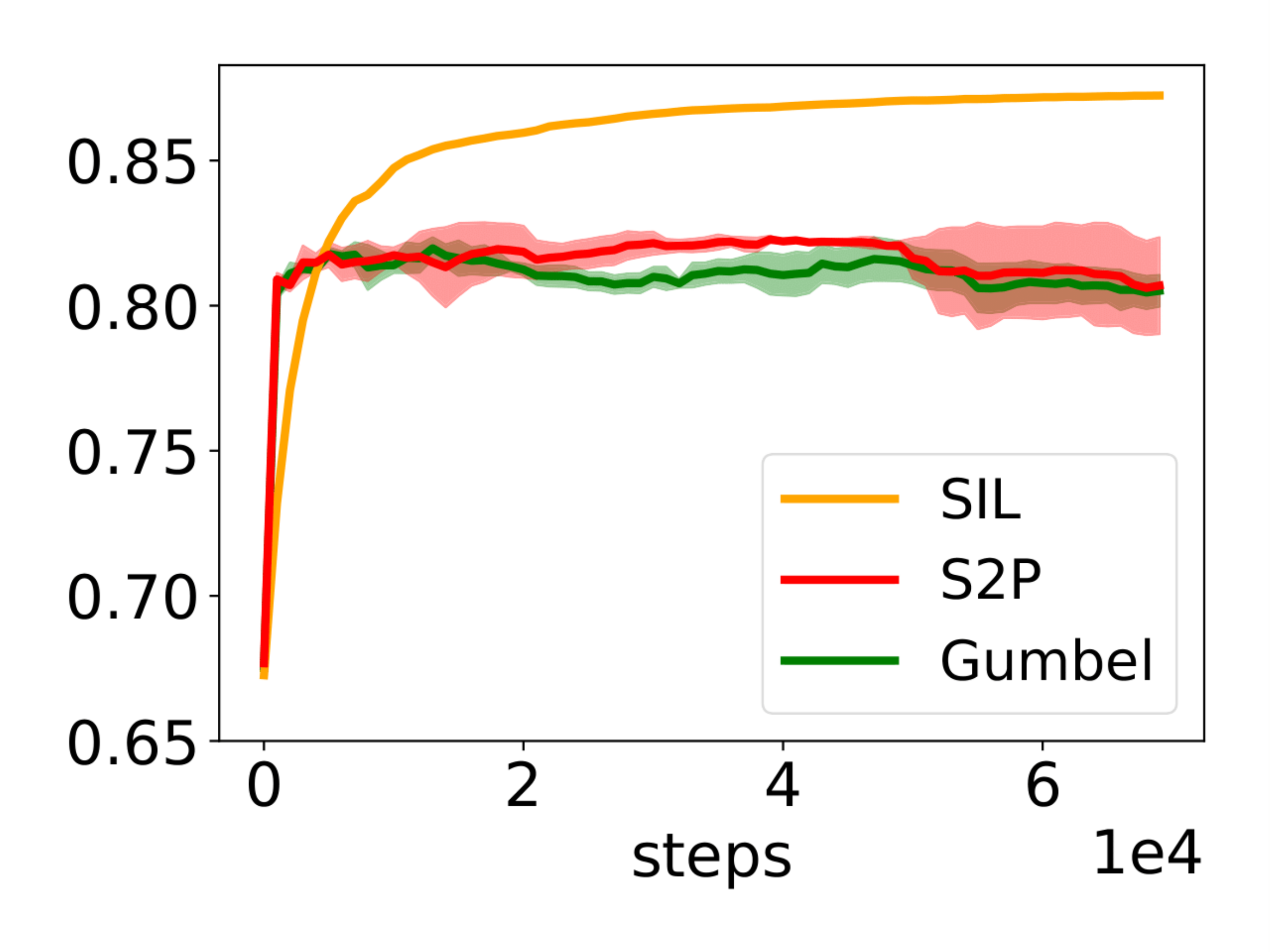}
          \vskip -0.5em
          \caption{Sender Language Score (Held-Out)}
        \end{subfigure}
        \hfill
        \begin{subfigure}[b]{0.32\columnwidth}
          \centering
          \includegraphics[width=\linewidth]{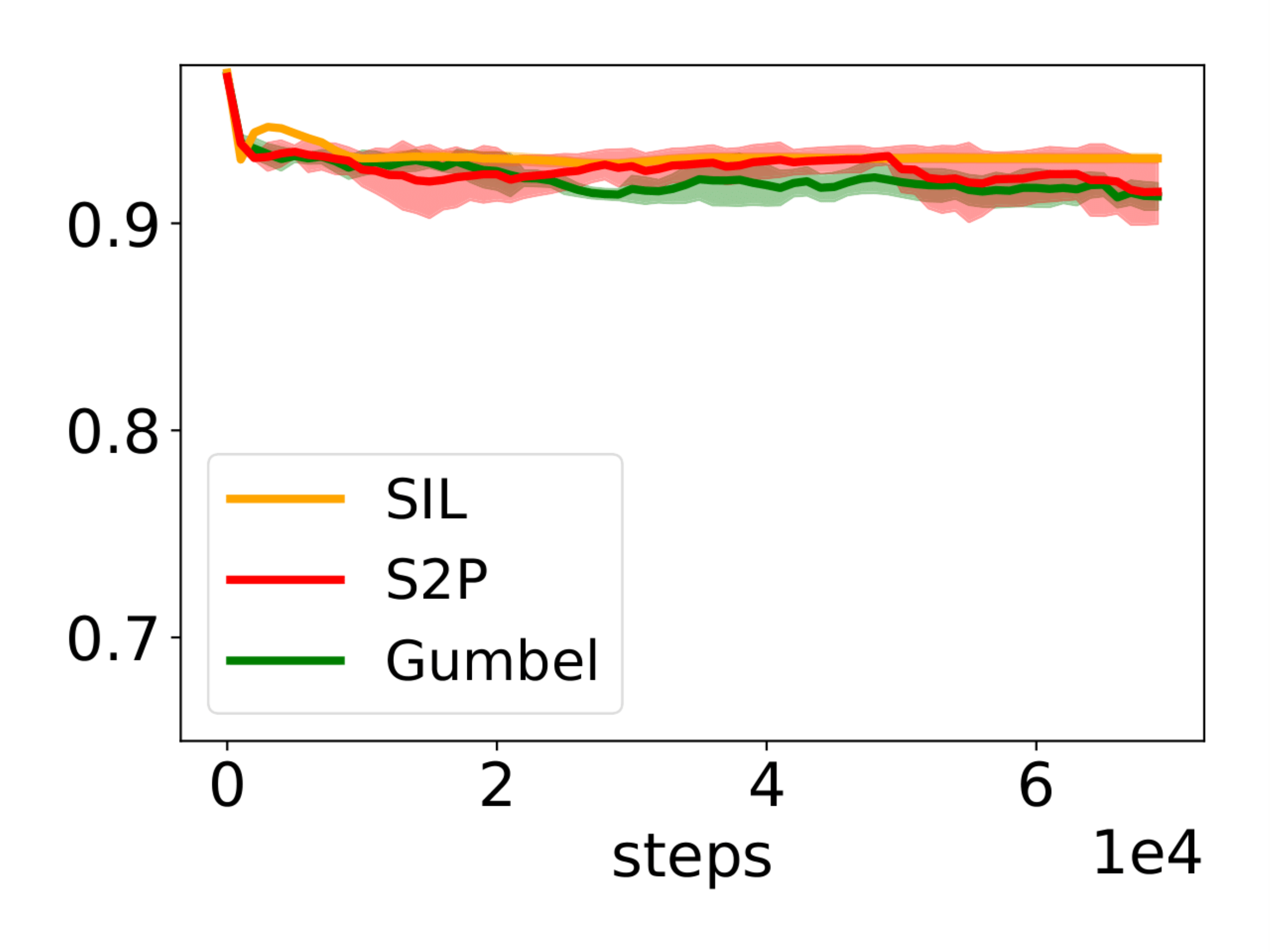}
          \vskip -0.5em
          \caption{Receiver Language Score (Held-Out)}
        \end{subfigure}
        \\
        \begin{subfigure}[b]{0.32\columnwidth}
          \centering
          \includegraphics[width=\linewidth]{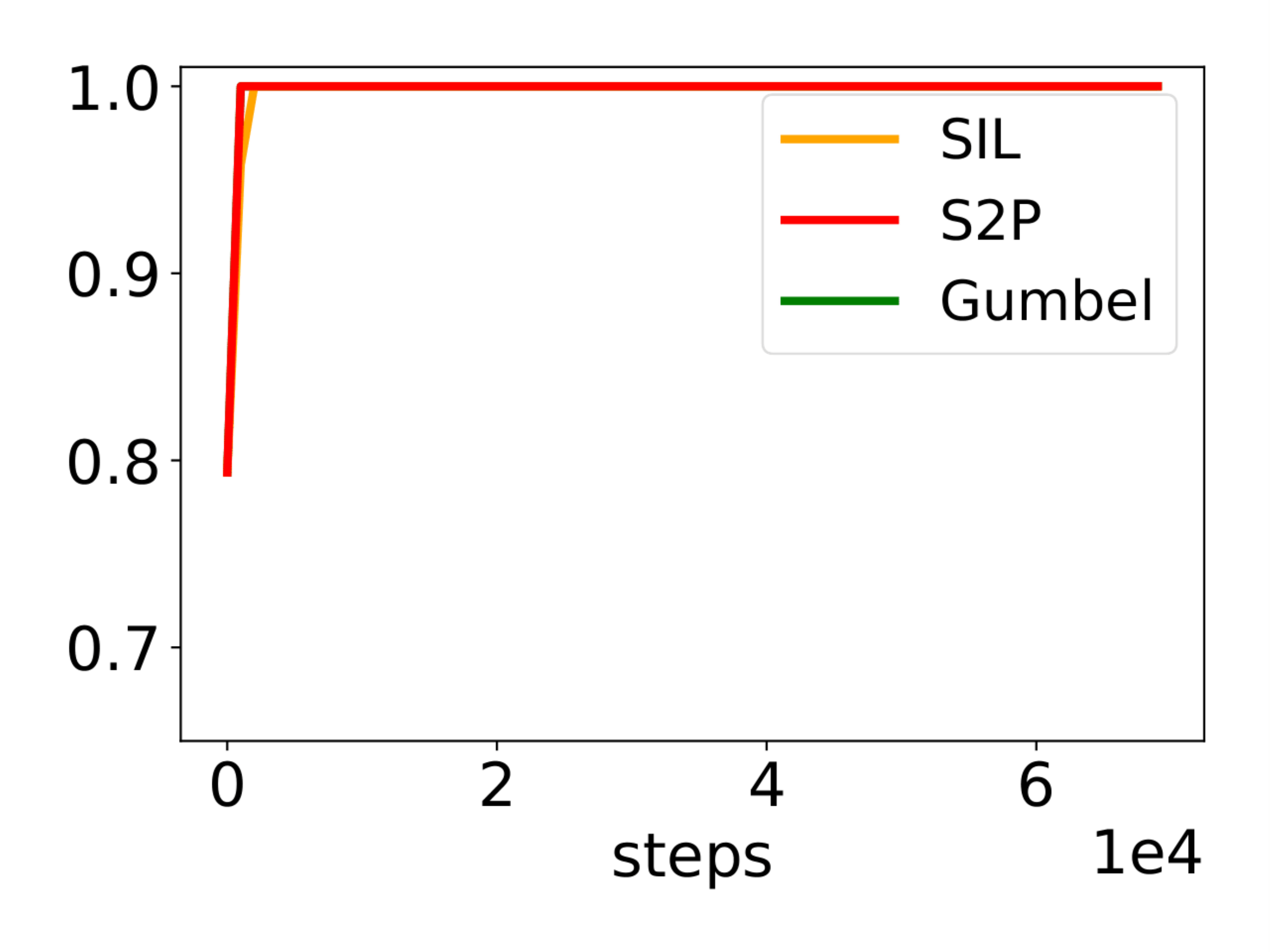} 
          \vskip -0.5em
          \caption{Task Score (Train)}
        \end{subfigure}
        \hfill
        \begin{subfigure}[b]{0.32\columnwidth}
          \centering
          \includegraphics[width=\linewidth]{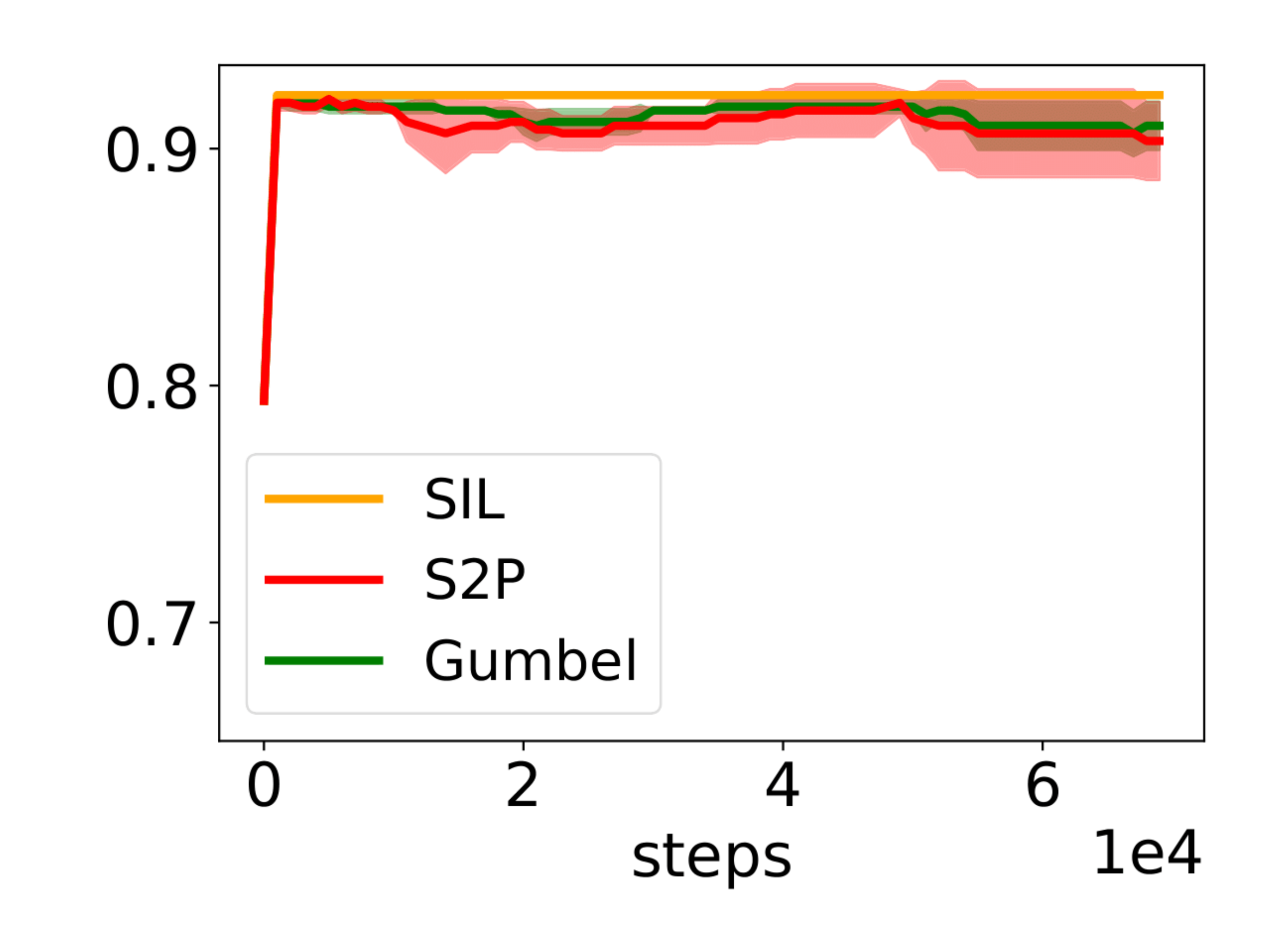} 
          \vskip -0.5em
          \caption{Sender Language Score (Train)}
        \end{subfigure}
        \hfill
        \begin{subfigure}[b]{0.32\columnwidth}
          \centering
          \includegraphics[width=\linewidth]{figures/lewis/g10/sp_listen_acc_gr_msg.pdf}
          \vskip -0.5em
          \caption{Receiver Language Score (Train)}
        \end{subfigure}
        \caption{Complete training curves for Task score and sender grounding in Lewis Game comparing \SIL vs baselines for $\tau=1$ on the held-out dataset (bottom), and the interactive training split (bottom). For \SIL we use $k_1=1000, k_2=k'_2=400$.}
        \label{fig:full_lewis_curve_1}
\end{figure}

We sweep over different Gumbel temperatures to assess the impact of exploration on language drift. We show the results with Gumbel temperature $\tau = 1, 10$ in  Fig \ref{fig:full_lewis_curve_1} and Fig \ref{fig:full_lewis_curve_10}. We observe that the baselines are very sensitive to Gumbel temperature: high temperature both decreases the language and tasks score. On the other side, \SILlong perform equally well on both temperatures and manage to maintain both task and language accuracies even with high temperature.

\subsection{Tracking Language Drift with Token Accuracy}
\begin{figure}[ht!]
        \centering
        \includegraphics[width=0.5\linewidth]{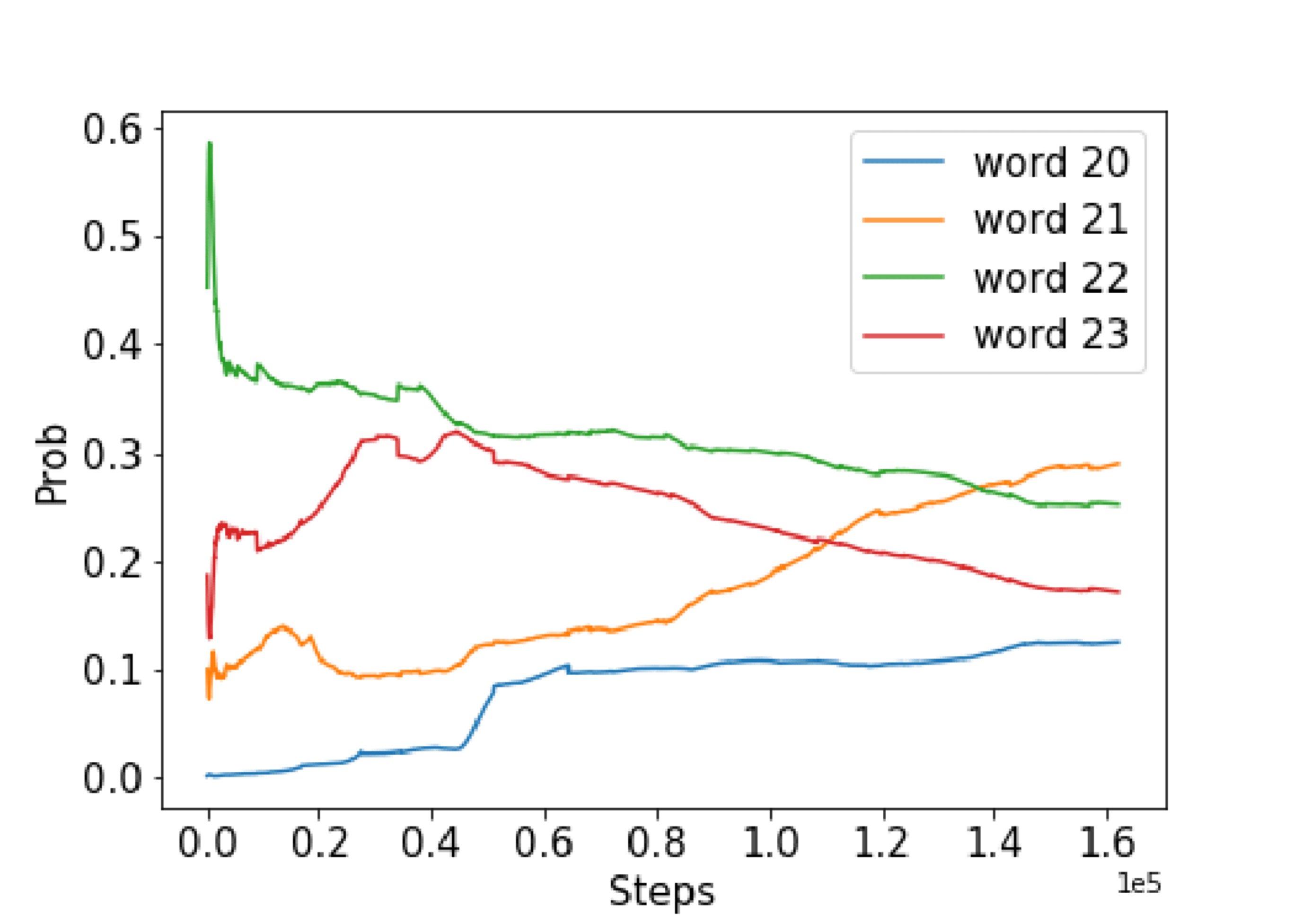}
        \caption{Change of conditional probability $s(w|c)$ where $c=22$ and $w=20, 21, 22, 23$. Following pretraining, $s(22|22)$ start with the highest probability. However, language drift gradually happens and eventually word 21 replaces the correct word 22.}
        \label{fig:token_evolution_lewis}
\end{figure}

To further visualize the language drift in Lewis game, we focus on the evolution of on the probability of speaking different word when facing the same concept. Formally, we track the change of conditional probability $s(w | c)$. The result is in Figure~\ref{fig:token_evolution_lewis}. 

\newpage
\section{Translation Game}
\label{appendix_translation_game}

\begin{table*}[ht!]
\small
\centering
\caption{Translation Game Results. The checkmark in ``ref len" means the method use reference length to constrain the output during training/testing. $\uparrow$ means higher the better and vice versa. Our results are averaged over 5 seeds,  and reported values are extracted for the best BLEU(De) score during training. We here use a Gumbel temperature of 0.5.}
\vskip 1em
\begin{tabular}{lll|cccc}
\toprule
\multicolumn{2}{l}{Method} & ref len & \multicolumn{2}{c}{BLEU$\uparrow$}  & NLL$\downarrow$ & R1\%$\uparrow$ \\ 
    &      &    & De &  En & \\
\midrule
\multirow{3}{*}{\citet{lee2019countering}}  
    & Pretrained & N/A          & 16.3                        & 27.18                       & N/A                          & N/A                    \\
    & PG         & $\checkmark$ & 24.51                       & 12.38                       & N/A                          & N/A                    \\
    & PG+LM+G    & $\checkmark$ & 28.08                       & 24.75                       & N/A                          & N/A                    \\ \midrule
 \multirow{7}{*}{Ours} 
 & Pretrained      & N/A         & 15.68                & 29.39                       & 2.49                         & 21.9                   \\
 & Fix Sender      & N/A         & 22.02 $\pm$ 0.18         & 29.39                       & 2.49                         & 21.9                   \\ 
 & Gumbel          &             & 27.11 $\pm$ 0.14         & 14.5 $\pm$ 0.83    & 5.33 $\pm$ 0.39           &9.7 $\pm$ 1.2   \\
 & Gumbel       & $\checkmark$   & 26.94 $\pm$ 0.20      & 23.41$\pm$ 0.50   & 5.04 $\pm$ 0.01            & 18.9 $\pm$ 0.8                   \\
 & S2P($\alpha=0.1$) &           & 27.43$\pm$ 0.36             & 19.16 $\pm$ 0.63              & 4.05 $\pm$ 0.16             & 13.6 $\pm$ 0.7                  \\
 & S2P($\alpha=1$)   &           & 27.35$\pm$ 0.19             & 29.73 $\pm$ 0.15              & 2.59 $\pm$ 0.02           & \textbf{23.7 $\pm$ 0.7}                   \\
 & S2P($\alpha=5$)   &           & 24.64$\pm$ 0.16             & \textbf{30.84 $\pm$ 0.07}     & 2.51 $\pm$ 0.02            & 23.5 $\pm$ 0.5                   \\
 & NIL         &                 & \textbf{28.29$\pm$ 0.16}    & 29.4 $\pm$ 0.25               & \textbf{2.15 $\pm$ 0.12}   & 21.7 $\pm$ 0.2                \\ \bottomrule  
\end{tabular}
\label{table:core}
\end{table*}

\begin{figure*}[ht!]
      \begin{subfigure}[b]{0.24\linewidth}
         \centering
         \includegraphics[width=\linewidth]{figures/translation/NILvsS2P/BLEU_De.pdf}
         \caption{BLEU De (Task Score)}
     \end{subfigure}
     \hfill
     \begin{subfigure}[b]{0.24\linewidth}
         \centering
         \includegraphics[width=\linewidth]{figures/translation/NILvsS2P/BLEU_En.pdf}
         \caption{BLEU En}
     \end{subfigure}
    \hspace{-0.5em}%
     \begin{subfigure}[b]{0.24\linewidth}
         \centering
         \includegraphics[width=\linewidth]{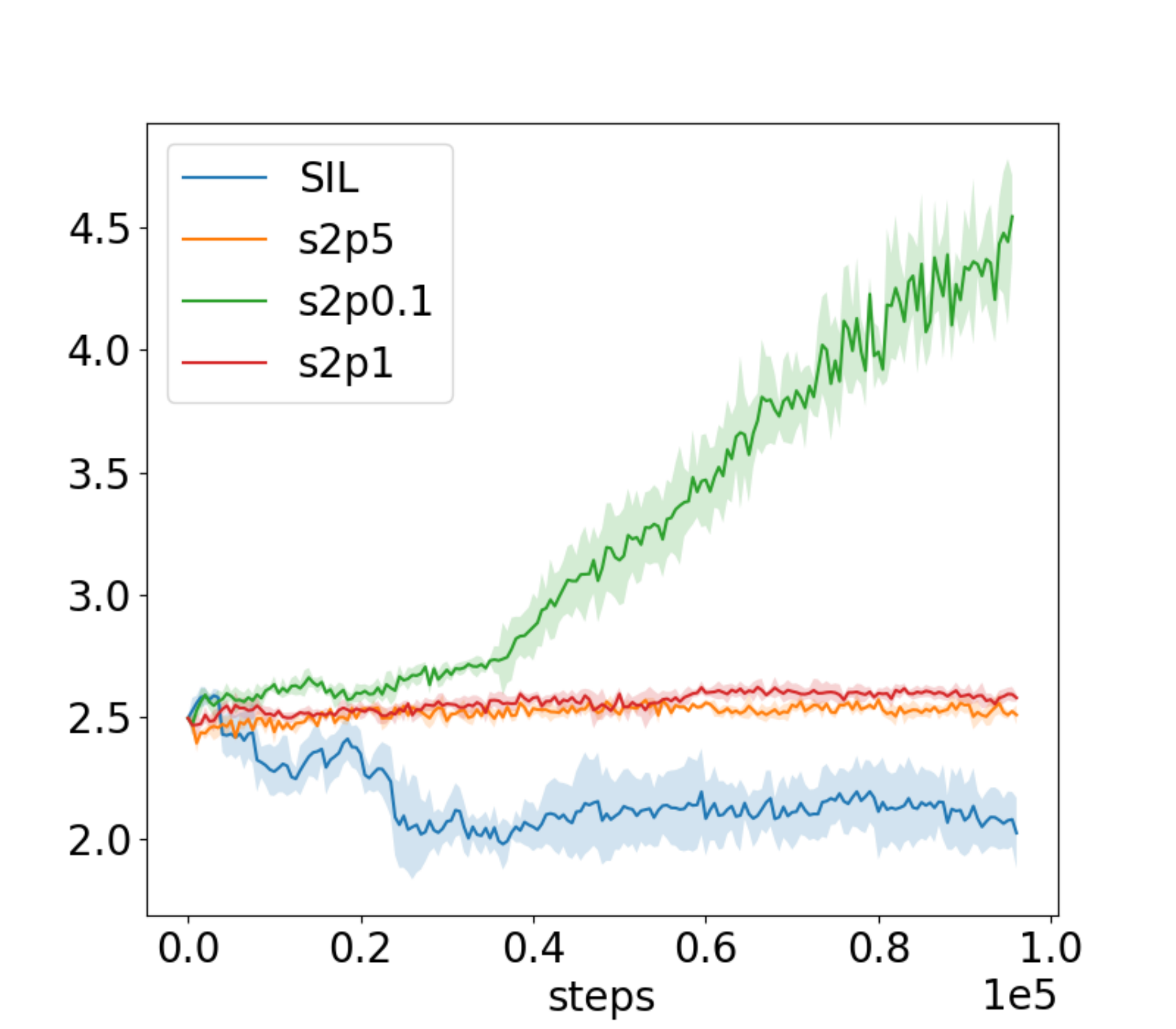}
         \caption{NLL}
     \end{subfigure}
     \hfill
    \begin{subfigure}[b]{0.24\linewidth}
         \centering
         \includegraphics[width=\linewidth]{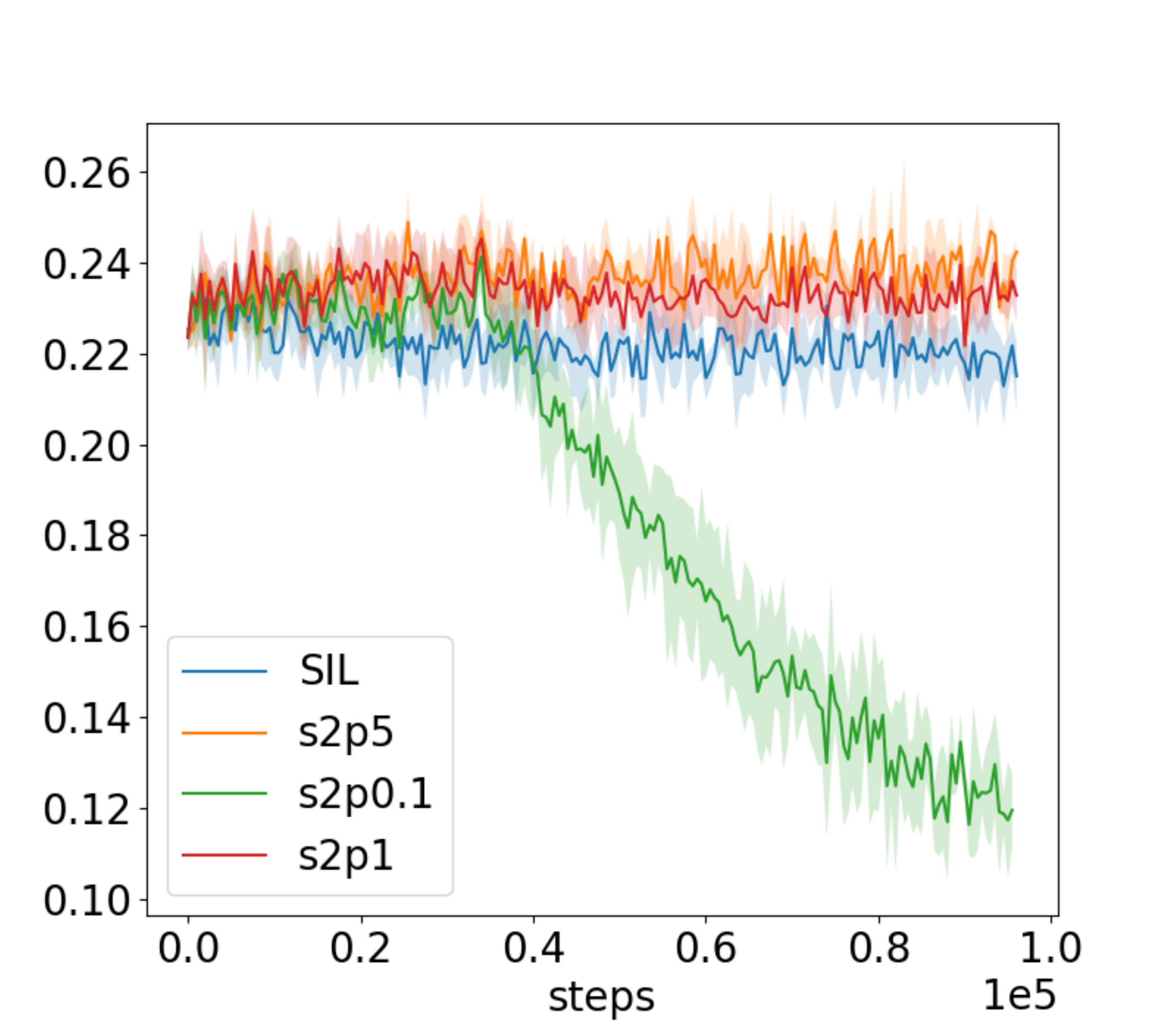}
         \caption{R1}
     \end{subfigure}
     \caption{S2P has a trade-off between the task score and the language score while \SIL is consistently high with both metrics.}
     \label{fig:s2p_vs_sil_full}
\end{figure*}

\subsection{Data Preprocessing}
We use Moses to tokenize the text~\citep{koehn2007moses} and we learn byte-pair-encoding~\citep{sennrich2016neural} from Multi30K~\citep{elliott2016multi30k} with all language. Then we apply the same BPE to different dataset. Our vocab size for En, Fr, De is 11552, 13331, and 12124.

\subsection{Model Details and Hyperparameters}
\label{sec:appendix_translation_model_details}
The model is a standard seq2seq translation model with attention~\citep{bahdanau2014neural}. Both encoder and decoder have a single-layer GRU~\citep{cho2014learning} with hidden size 256. The embedding size is 256. There is a dropout after embedding layers for both encoder and decoder
For decoder at each step, we concatenate the input and the attention context from last step. 

\paragraph{Pretraining} For Fr-En agent, we use dropout ratio 0.2, batch size 2000 and learning rate 3e-4. We employ a linear learning rate schedule with the anneal steps of 500k. The minimum learning rate is 1e-5. We use Adam optimizer~\citep{kingma2014adam} with $\beta=(0.9, 0.98)$. We employ a gradient clipping of 0.1. For En-De, the dropout ratio is 0.3. We obtain a BLEU score of 32.17 for Fr-En, and 20.2 for En-De on the IWSLT test dataset~\citep{cettolo2012wit3}. 

\paragraph{Finetuning}
During finetuning, we use batch size 1024 and learning rate 1e-5 with no schedule. The maximum decoding length is 40 and minimum decoding length is 3. For iterated learning, we use $k_1=4000$, $k_2=200$ and $k'_2=300$. We set Gumbel temperature to be $5$. We use greedy sample from teacher speaker for imitation.

\subsection{Language Model and Image Ranker Details}
\label{sec:appendix_translation_lm_ranker}
Our language model is a single-layer LSTM~\citep{hochreiter1997long} with hidden size 512 and embedding size 512. We use Adam and learning rate of 3e-4. We use a batch size of 256 and a linear schedule with 30k anneal steps. The language model is trained with captions from MSCOCO~\citep{lin2014microsoft}. For the image ranker, we use the pretrained ResNet-152~\citep{he2016deep} to extract the image features. We use a GRU~\citep{cho2014learning} with hidden size 1024 and embedding size 300. We use a batch size of 256 and use VSE loss~\citep{faghri2017vse}. We use Adam with learning rate of 3e-4 and a schedule with 3000 anneal steps~\citep{kingma2014adam}.

\subsection{Language Statistics}
\label{sec:language_statistics}

\begin{figure*}[ht!]
    \centering
        \begin{subfigure}[b]{0.32\columnwidth}
         \centering
         \includegraphics[width=\linewidth]{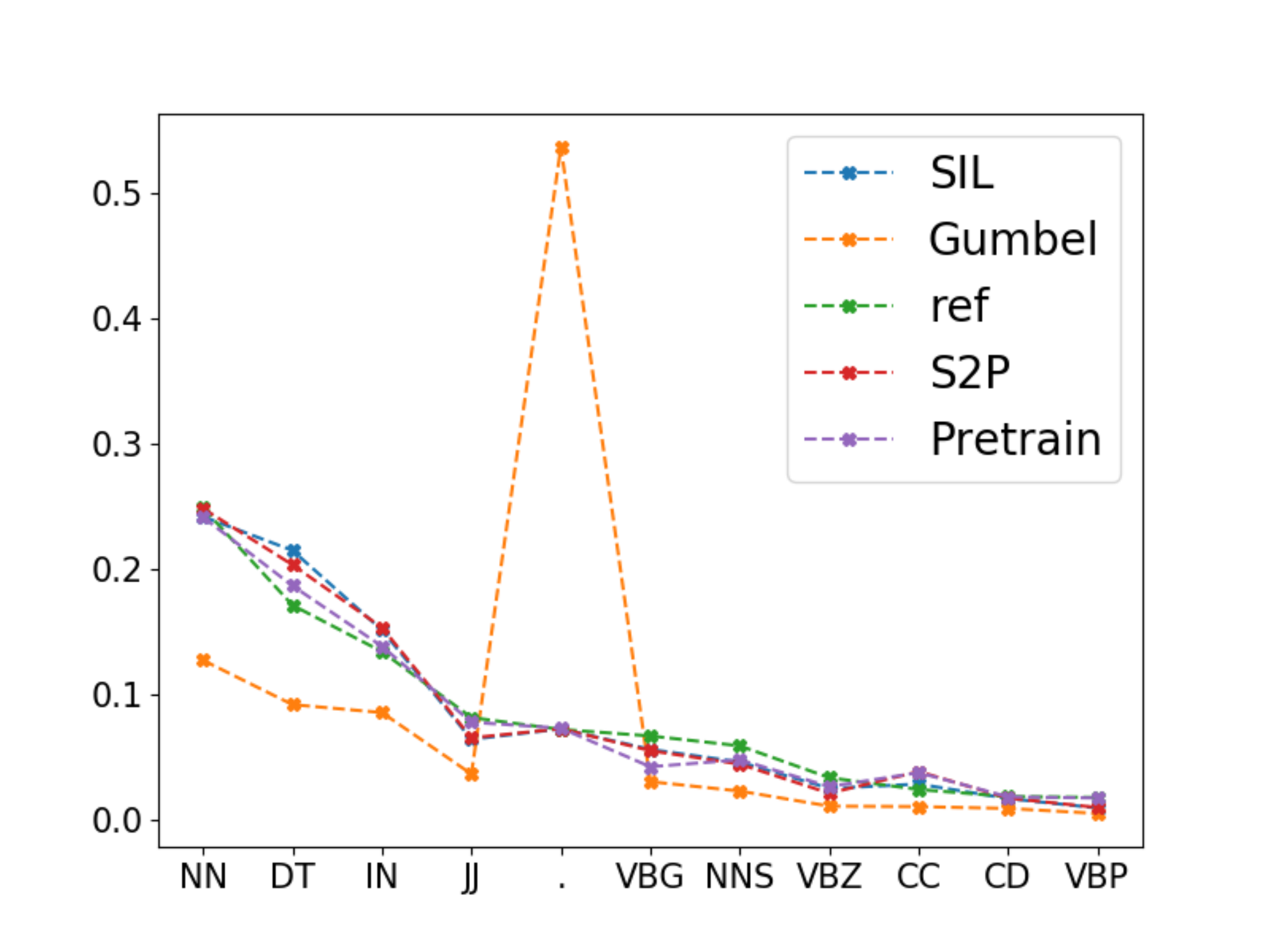}
         \caption{POS tag distribution.}
         \label{fig:pos_tag}
     \end{subfigure}
    \hfill
        \begin{subfigure}[b]{0.32\columnwidth}
         \centering
         \includegraphics[width=\linewidth]{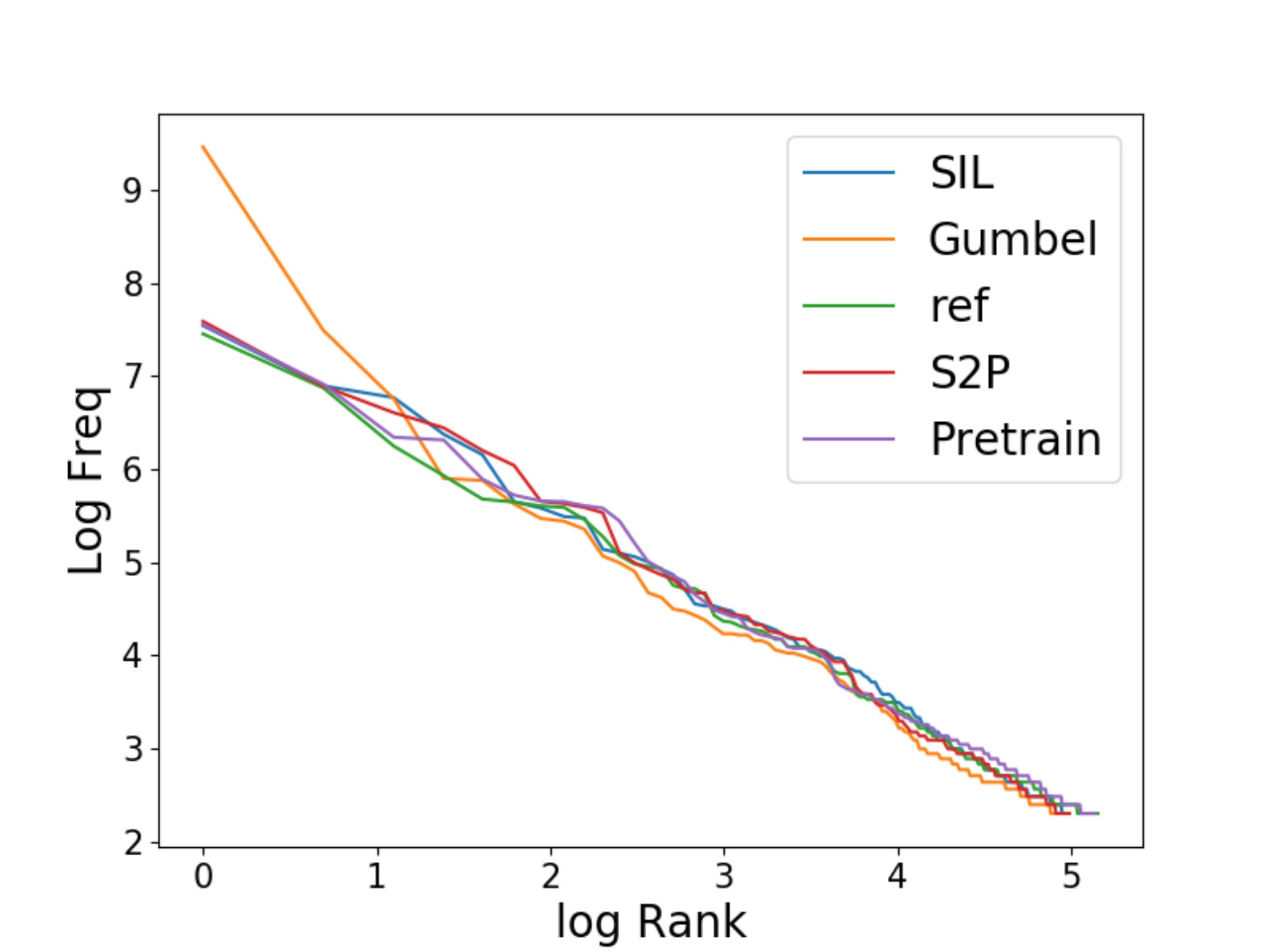}
         \caption{Word Frequency Analysis}
         \label{fig:freq}
    \end{subfigure}
    \hfill
        \begin{subfigure}[b]{0.32\columnwidth}
         \centering
         \includegraphics[width=\linewidth]{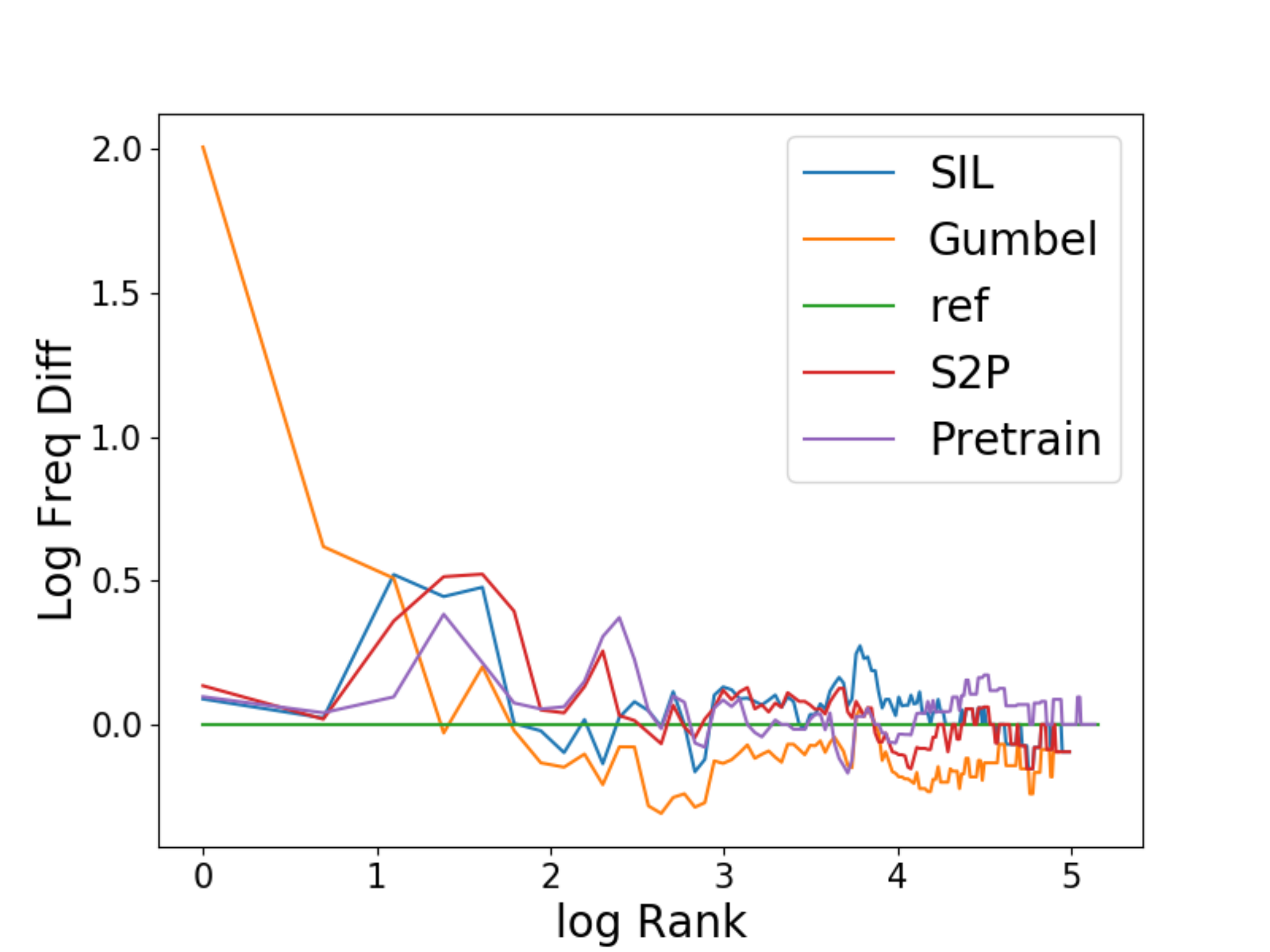}
         \caption{Difference of Log of Word Frequency}
         \label{fig:freq_diff}
    \end{subfigure}
    \caption{Language statistics on samples from different method.}
    \label{fig:sample_analysis}
\end{figure*}
We here compute several linguistic statistics on the generated samples to assess language quality.

\paragraph{POS Tag Distribution}
We compute the Part-of-Speech Tag (POS Tag~\citep{marcus1993building}) distribution by counting the frequency of POS tags and normalize it. The POS tag are sorted according to their frequencies in the reference, and we pick the 11 most common POS tag for visualization, which are:

\vspace{-0.6em}
\begin{itemize}
    \setlength\itemsep{0em}  
   \item NN	Noun, singular or mass
\item DT	Determiner
\item IN	Preposition or subordinating conjunction
\item JJ	Adjective
\item VBG	Verb, gerund or present participle
\item NNS	Noun, plural
\item VBZ	Verb, 3rd person singular present
\item CC	Coordinating conjunction
\item CD	Cardinal number
\end{itemize}

The results are shown in Figure~\ref{fig:pos_tag}. The peak on ``period" show that Gumbel has tendency of repeating periods at the end of sentences. However, we observe that both S2P and 

\paragraph{Word Frequency}
For each generated text, we sort the frequency of the words and plot the log of frequency vs. log of rank. We set a minimum frequency of 50 to exclude long tail results. The result is in Figure~\ref{fig:freq}. 

\paragraph{Word Frequency Difference}
To further visualize the difference between generated samples and reference, we plot the difference between their log of word frequencies in Figure~\ref{fig:freq_diff}.

\section{Human Evaluation}
We here assess whether our language drift evaluation correlates with human judgement. To do so, we performed a human evaluation with two pairwise comparison tasks.
\begin{itemize}
    \item In Task1, the participant picks the best English semantic translation while observing the French sentence.
    \item In Task2, the participant picks the best English translation from two candidates.
\end{itemize}
Thus, the participants are likely to rank captions mainly by their syntax/grammar quality in Task2, whereas they would also consider semantics in Task1, allowing us to partially disentangle structural and semantic drift.

For each task, we use the validation data from Multi30K (1013 French captions) and generate 4 English sentences for each French caption from the Pretrain, Gumbel, S2P, and SIL. We also retrieved the ground-truth human English caption. We then build the test by randomly sampling two out of five English captions. We gathered 22 people, and we collect about 638 pairwise comparisons for Task2 and 315 pairwise comparisons for Task1. We present the result in Table~\ref{tab:human_eval_task1} and Table~\ref{tab:human_eval_task2}. I also include the binomial statistical test result where the null hypothesis is \emph{methods are the same}, and the alternative hypothesis is \emph{one method is better than the other one}.


Unsurprisingly, we observe that the Human samples are always preferred over generated sentences. Similarly, Gumbel is substantially less preferred than other models in both settings.

In Task 1(French provided), human users always preferred S2P and SIL over pretrained models with a higher win ratio. 
Oh the other hand when French is not provided, the human users prefer the pretrain models over S2P and SIL. We argue that while the pretrained model keeps generating gramartically correct sentences, its translation effectiveness is worse than both S2P and SIL since these two models go through the interactive learning to adapt to new domain.

Finally, SIL seems to be preferred over S2P by a small margin in both tasks.
However, our current ranking is not conclusive, since we can see the significance level of comparisons among Pretrain, S2P, and SIL is not smaller enough to reject null hypothesis, especially in task 1 where we have less data points. In the future we plan to have a larger scale human evaluation to further differentiate these methods.



\begin{table}[!ht]
 \caption{The Win-Ratio Results. The number in row $X$ and column $Y$ is the empiric ratio that method $X$ beats method $Y$ according collected human pairwise preferences. We perform a naive ranking by the row-sum of win-ratios of each method.  We also provide the corresponding P-values under each table. The null hypothesis is \emph{two methods are the same}, while the alternative hypothesis is \emph{two methods are different.} }
       \centering
    \begin{minipage}{.49\linewidth}
      \caption{With French Sentences}\label{}
\begin{tabular}{llllll}
\toprule
         & Gumbel & Pretrain & S2P & SIL & Human \\ \midrule
Gumbel   & 0      & 0.25     & 0.15  & 0.12  & 0      \\
Pretrain & 0.75   & 0        & 0.4   & 0.4  & 0.13  \\
S2P      & 0.84   & 0.6      & 0     & 0.38  & 0.21   \\
SIL      & 0.88   & 0.6     & 0.63  & 0     & 0.22    \\
Human      & 1      & 0.87     & 0.79  & 0.77  & 0     \\ \midrule
Ranking  & \multicolumn{5}{l}{Human(3.4), SIL(2.3), S2P(2.0), Pretrain(1.7), Gumbel(0.5)}                     \\ \bottomrule
\multicolumn{6}{c}{P-values} \\ \midrule
         & Gumbel & Pretrain & S2P & SIL & Human   \\ \midrule
Gumbel   & -            & $<10^{-2}$   & $<10^{-2}$  & $<10^{-2}$  & $<10^{-2}$      \\
Pretrain & $<10^{-2}$   & -            & $0.18$  & $0.21$  & $<10^{-2}$  \\
S2P      & $<10^{-2}$   & $0.18$       & -       & $0.15$  & $<10^{-2}$  \\
SIL      & $<10^{-2}$   & $0.21$        & $0.15$  & -       & $<10^{-2}$    \\
Human    & $<10^{-2}$   & $<10^{-2}$   & $<10^{-2}$  & $<10^{-2}$  & -      \\ 
\bottomrule
\end{tabular}
\label{tab:human_eval_task1}
    \end{minipage}%
    \\
    \centering
    \begin{minipage}{.49\linewidth}
        \caption{Without French Sentences}
\begin{tabular}{llllll}
\toprule
         & Gumbel & Pretrain & S2P & SIL & Human \\ \midrule
Gumbel   & 0      & 0.16  & 0.12  & 0.13  & 0.02    \\
Pretrain & 0.84   & 0     & 0.69  & 0.59  & 0.15 \\
S2P      & 0.88   & 0.31  & 0     & 0.38  & 0.05   \\
SIL      & 0.86   & 0.41  & 0.62  & 0     & 0.01    \\
Human    & 0.98   & 0.85  & 0.95  & 0.98  & 0     \\ \midrule
Ranking  & \multicolumn{5}{l}{Human(3.8), Pretrain(2.3), SIL(1.9), S2P(1.6), Gumbel(0.4)}                      \\ \midrule
\multicolumn{6}{c}{P-values} \\ \midrule
         & Gumbel & Pretrain & S2P & SIL & Human   \\ \midrule
Gumbel   & -      & $<10^{-2}$     & $<10^{-2}$  & $<10^{-2}$  & $<10^{-2}$      \\
Pretrain & $<10^{-2}$   & -       & $<10^{-2}$  & $0.08$  & $<10^{-2}$  \\
S2P      & $<10^{-2}$   & $<10^{-2}$  & -       & $0.06$  & $<10^{-2}$  \\
SIL      & $<10^{-2}$   & $0.08$     & $0.06$   & -     & $<10^{-2}$    \\
Human    & $<10^{-2}$   & $<10^{-2}$   & $<10^{-2}$  & $<10^{-2}$  & -      \\ 
\bottomrule
\end{tabular}
\label{tab:human_eval_task2}
    \end{minipage} 
\end{table}

\section{Samples}
We list more samples from the Multi30k dataset with different baselines, i.e., Pretrain, Gumbel, S2P($\alpha=1$. The Gumbel temperature is set to 0.5. The complete samples can be found in our code.
{
\small

ref     : a female playing a song on her violin . \\
Pretrain: a woman playing a piece on her violin . \\ 
Gumbel  : a woman playing a piece on his violin . . . . . . . . . . . . . \\ 
S2P     : a woman playing a piece on his violin . \\ 
SIL     : a woman playing a piece on his violin . \\

ref     : a cute baby is smiling at another child . \\
Pretrain: a nice baby smiles at another child .\\
Gumbel  : a nice baby smiles of another child . . . . . . . . . . \\
S2P     : a nice baby smiles at another child . \\
SIL     : a beautiful baby smiles smiles at another child . \\

ref     : a man drives an old-fashioned red race car . \\
Pretrain: a man conducted an old race car .\\
Gumbel  : a man drives a old race of red race . . . .\\
S2P     : a man drives an old of the red race .\\
SIL     : a man drives a old race of the red race .\\

ref     : a man in a harness climbing a rock wall \\
Pretrain: a man named after a rock man . \\
Gumbel  : a man thththththththdeacdeaacc. of th. . . . . . . \\
S2P     : a man 's being a kind of a kind of a kind . \\
SIL     : a man that the datawall of the datad. \\

ref     : a man and woman fishing at the beach . \\
Pretrain: a man and a woman is a woman . \\
Gumbel  : a man and a woman thaccbeach the beach . . . . . . . . . . \\
S2P     : a man and a woman is in the beach . \\
SIL     : a man and a woman that 's going to the beach . \\

ref     : a man cooking burgers on a black grill . \\ 
Pretrain: a man making the meets on a black slick of a black slick . \\
Gumbel  : a man doing it of on a black barbecue . . . . . . . . . . . . . . . .\\
S2P     : a man doing the kind on a black barbecue . \\
SIL     : a man doing the datadon a black barbecue . \\

ref     : little boy in cami crawling on brown floor \\
Pretrain: a little boy in combination with brown soil . \\
Gumbel  : a small boy combincombinaccon a brown floor . . . brown . . . . . . . . . \\
S2P     : a small boy combining the kind of brown floor . \\ 
SIL     : a small boy in the combination of on a brown floor .\\

ref     : dog in plants crouches to look at camera . \\
Pretrain: a dog in the middle of plants are coming to look at the goal . \\
Gumbel  : a dog in middle of of of of thlooking at looking at objeobje. . . . . . . . . . . . . . . . . . . \\
S2P     : a dog in the middle of the plants to watch objective . \\
SIL     : a dog at the middle of plants are going to look at the objective . \\

ref     : men wearing blue uniforms sit on a bus . \\
Pretrain: men wearing black uniforms are sitting in a bus . \\
Gumbel  : men wearing blue uniforms sitting in a bus . . . . . . . \\
S2P     : men wearing blue uniforms sitting in a bus . \\
SIL     : men wearing blue uniforms are sitting in a bus . \\

ref     : a group of scottish officers doing a demonstration . \\
Pretrain: a scottish officers group is doing a demonstration . \\
Gumbel  : a group of officers scottish doing a dedemonstration . . . . \\
S2P     : a group of officers scottish doing a demonstration . \\
SIL     : a group of officers scottish doing a demo . \\

ref     : the brown dog is wearing a black collar .  \\
Pretrain: the brown dog is wearing a black collar . \\
Gumbel  : the brown dog carries a black collar . . . . . . . \\
S2P     : the brown dog carries a black collar . \\
SIL     : the brown dog is wearing a black collar . \\

ref     : twp children dig holes in the dirt . \\
Pretrain: two children are going to dig holes in the earth . \\
Gumbel  : two children dig holes in the planplanplanplan. . . . . . . . \\
S2P     : two children are going holes in the dirt . \\
SIL     : two children dig holes in the earth . \\

ref     : the skiers are in front of the lodge . \\
Pretrain: the health are in front of the bed . \\
Gumbel  : the ththare ahead the thth. . . . . . . \\
S2P     : the health are front of the whole . \\
SIL     : the dataare are ahead of the datad. \\

ref     : a seated man is working with his hands . \\
Pretrain: a man sitting working with his hands . \\
Gumbel  : a man sitting working with his hands . . . . . . . . . \\
S2P     : a man sitting working with his hands . \\
SIL     : a man sitting working with its hands . \\

ref     : a young girl is swimming in a pool . \\
Pretrain: a girl swimming in a swimming pool . \\
Gumbel  : a young girl swimming in a pool . . . . . . . . . . \\
S2P     : a young girl swimming in a pool . \\
SIL     : a young girl swimming in a pool . \\

ref     : a small blond girl is holding a sandwich . \\
Pretrain: a little girl who is a sandwich . \\
Gumbel  : a yedegirl holding a sandwich . . . . \\
S2P     : a small 1girl holding a sandwich . \\
SIL     : a small 1girl holding a sandwich . \\

ref     : two women look out at many houses below . \\
Pretrain: two women are looking at many of the houses in the computer . \\
Gumbel  : two women looking many of many houses in itdeacede. . . . . . . . \\
S2P     : two women looking at many houses in the kind . \\
SIL     : two women looking at many houses in the data. \\

ref     : a person is hang gliding in the ocean . \\
Pretrain: ( wind up instead of making a little bit of the board ) a person who is the board of the sailing . \\
Gumbel  : ( cdthinplace of acacc) a person does thacthof th-acin the ocean . . . . . . . . . . . . . . . . \\
S2P     : ( wind 's instead of a kind ) a person does the kind in the ocean . \\
SIL     : ( datadinstead of the input of the clinability ) a person does the board in the ocean . \\

ref     : a man in a green jacket is smiling . \\
Pretrain: a green man in the green man . \\
Gumbel  : a man jacket green smiles . . . . . . . . . . . . \\
S2P     : a man in jacket green smiles . \\
SIL     : a man in the green jacket smiles . \\

ref     : a young girl standing in a grassy field . \\
Pretrain: a girl standing in a meadow . \\
Gumbel  : a young girl standing in a gmeadow . . . . . . . . . \\
S2P     : a young girl standing in a meadow . \\
SIL     : a young girl standing in a meadow . \\  

}
    
\end{document}